\documentclass{article}

\usepackage{microtype}
\usepackage{graphicx}
\usepackage{booktabs} 
\usepackage{hyperref}

\usepackage[accepted]{icml2024}

\usepackage{mathtools}
\usepackage[utf8]{inputenc} 
\usepackage[T1]{fontenc}    
\usepackage{url}            
\usepackage{amsfonts}       
\usepackage{nicefrac}       
\usepackage{xcolor}         
\usepackage{colortbl}
\usepackage{subcaption}
\usepackage{enumitem}
\usepackage{multirow}
\usepackage{capt-of}
\usepackage{wrapfig}
\usepackage{makecell}
\usepackage{bm}
\usepackage{arydshln}
\usepackage{placeins}
\hypersetup{
    colorlinks = true,
    citecolor = blue,
    linkcolor = red,
    urlcolor = blue,
}
\usepackage{comment}
\usepackage{algorithm}
\usepackage{algorithmic}
\usepackage{eqparbox}
\usepackage{amsmath, amsfonts, amsmath, amssymb, amsthm, bm, bbm}
\usepackage{pifont} % for checkmark

\theoremstyle{plain}

\theoremstyle{definition}

\theoremstyle{remark}

\renewcommand{\algorithmiccomment}[1]{\hfill{ $\rhd$ #1}}

\newcommand{\overbar}[1]{\mkern 1.5mu\overline{\mkern-1.5mu#1\mkern-1.5mu}\mkern 1.5mu}

\usepackage{adjustbox}

\definecolor{gg}{gray}{0.92}
\newcolumntype{a}{>{\columncolor{gg}}c}
\definecolor{figblue}{RGB}{47, 85, 151}
\definecolor{figpurple}{RGB}{112, 48, 160}

\usepackage[capitalize,noabbrev]{cleveref}
\usepackage[hang,flushmargin]{footmisc} % remove indent from the footnote

\usepackage[textsize=tiny]{todonotes}

\icmltitlerunning{Generative Modeling on Manifolds Through Mixture of Riemannian Diffusion Processes}

\begin{document}

\twocolumn[
\icmltitle{Generative Modeling on Manifolds Through \\ Mixture of Riemannian Diffusion Processes}

\begin{icmlauthorlist}
\icmlauthor{Jaehyeong Jo}{kaist}
\icmlauthor{Sung Ju Hwang}{kaist,deepauto}
\end{icmlauthorlist}

\icmlaffiliation{kaist}{Korea Advanced Institute of Science and Technology (KAIST)}
\icmlaffiliation{deepauto}{DeepAuto.ai}

\icmlcorrespondingauthor{Jaehyeong Jo}{harryjo97@kaist.ac.kr}
\icmlcorrespondingauthor{Sung Ju Hwang}{sjhwang82@kaist.ac.kr}

\icmlkeywords{Geometric Deep Learning, Generative Models}
\vskip 0.3in
]
\printAffiliationsAndNotice{} % otherwise use the standard text.

\begin{abstract}
Learning the distribution of data on Riemannian manifolds is crucial for modeling data from non-Euclidean space, which is required by many applications in diverse scientific fields. Yet, existing generative models on manifolds suffer from expensive divergence computation or rely on approximations of heat kernel. These limitations restrict their applicability to simple geometries and hinder scalability to high dimensions. In this work, we introduce the Riemannian Diffusion Mixture, a principled framework for building a generative diffusion process on manifolds. Instead of following the denoising approach of previous diffusion models, we construct a diffusion process using a mixture of bridge processes derived on general manifolds without requiring heat kernel estimations.
We develop a geometric understanding of the mixture process, deriving the drift as a weighted mean of tangent directions to the data points that guides the process toward the data distribution.
We further propose a scalable training objective for learning the mixture process that readily applies to general manifolds. Our method achieves superior performance on diverse manifolds with dramatically reduced number of in-training simulation steps for general manifolds.\protect\footnotemark
\end{abstract}
\section{Introduction}
Deep generative models have shown great success in learning the distributions of the data represented in Euclidean space, e.g., images and text. While the focus of the previous works has been biased toward data in the Euclidean space, modeling the distribution of the data that naturally resides in manifolds with specific geometry has been underexplored, while they are required for wide application: For example, the earth and climate science data~\citep{karpatne2018geoscience, mathieu2020rcnf} lives in the sphere, whereas protein structures~\citep{jumper2021highly, watson2022broadly} and robotic movements~\citep{simeonov2022neural} are best represented by the group SE(3), and 3D computer graphics shapes~\citep{hoppe1992surface} can be identified as a general closed manifold. However, previous generative methods are ill-suited for modeling these data as they do not take into consideration the specific geometry describing the data space and may assign a non-zero probability to regions outside the desired space.

\footnotetext{Code: \href{https://github.com/harryjo97/riemannian-diffusion-mixture}{github.com/harryjo97/riemannian-diffusion-mixture}}

Recent works~\citep{debortoli2022rsgm, huang2022rdm} extend the diffusion generative framework to the Riemannian manifolds that learn to reverse the noising process, similar to Euclidean diffusion models. Although diffusion models have been shown to successfully model the distribution on simple manifolds, e.g., sphere and torus, they have difficulty in training since the score matching objective either relies on an imprecise approximation of the intractable heat kernel that degrades the performance or requires the computation of the divergence which is computationally expensive and scales poorly to high dimensions. 
In addition, previous diffusion models are geometrically not intuitive as their generative processes are parameterized by the score function which does not provide explicit geometric interpretation.

On the other hand, continuous normalizing flow (CNF) models on manifold~\citep{mathieu2020rcnf, rozen2021moser, ben-hamu2022cnfm, chen2024rfm} aim to learn the continuous-time flow by parameterizing the vector field. While CNF models leverage deterministic processes and alleviate the challenges of leveraging Brownian motion, most of the CNF models require computation of divergence during training that cannot even scale to moderately high dimensions and further cannot be readily adapted to general geometries. Even though several works~\citep{rozen2021moser, ben-hamu2022cnfm, chen2024rfm} proposed simulation-free methods on simple manifolds, they still require in-training simulation for general manifolds which necessitates a large number of steps to obtain accurate trajectories for the deterministic process.

In this work, we present Riemannian Diffusion Mixture, a novel generation framework for learning a diffusion process on Riemannian manifolds based on a geometric perspective.
We build upon the diffusion mixture representation~\citep{peluchetti2021mixture, liu2023bridge}, constructing a diffusion process directly on the manifold as a mixture of bridge processes, i.e., diffusion process conditioned to endpoints, without the need for heat kernel estimation.
We show that by designing the drift of a diffusion process as a weighted mean of tangent vectors to the data points, the resulting process is guided to the data distribution and yields a prediction of the final result as the most probable endpoint.
We further derive a scalable training objective, namely the two-way bridge matching, based on simple regression on the drifts of the diffusion processes that do not require computing divergence.
We establish a theoretical background for the diffusion mixture framework on the Riemannian setting that is readily applicable to general manifolds and show that the previous CNF model is a special case of our framework.

We experimentally validate our approach on diverse manifolds of both real-world and synthetic datasets, on which our method outperforms or is on par with the state-of-the-art baselines. We demonstrate that ours can scale to high dimensions while allowing significantly faster training compared to the previous diffusion models relying on score matching.
Especially on general manifolds, our method shows superior performance with dramatically reduced in-training simulation steps, using only 5\% of the steps compared to CNF model. We summarize our main contributions as follows:

\begin{itemize}[itemsep=0.5mm, parsep=3pt, leftmargin=*]
    \item We propose a principled framework for building generation processes on general manifolds as a mixture of bridge processes that does not require estimation of heat kernel.
    \item We present a geometric design for the drift of the diffusion process as a weighted mean of tangent directions on manifolds that guides the process to the target distribution, and introduce an efficient training objective readily applicable to general manifolds.
    \item Our method achieves superior performance on diverse manifolds, and we empirically show that ours can scale to higher dimensions with significantly faster training compared to previous diffusion models.
    \item Especially on general manifolds, our approach outperforms CNF model with greatly reduced in-training simulation steps, demonstrating the necessity of stochasticity.
\end{itemize}
\section{Background}
In this section, we introduce basic concepts of Riemannian manifolds and diffusion processes defined on manifolds.

\paragraph{Riemannian Manifold}
We consider complete, orientable, connected, and boundaryless Riemannian manifolds $\mathcal{M}$ equipped with Riemannian metric $g$ that defines the inner product of tangent vectors. $T_x\mathcal{M}$ denotes the tangent space at point $x\in\mathcal{M}$ and $\|\eta\|_{\mathcal{M}}$ denotes the norm of the tangent vector $\eta\in T_x\mathcal{M}$. 
For smooth function $f:\mathcal{M}\rightarrow\mathbb{R}$, $\nabla f(x)\in T_x\mathcal{M}$ denotes the Riemannian gradient, $\text{div}(v)$ denotes the Riemannian divergence for the smooth vector field $v:\mathcal{M}\rightarrow T_x\mathcal{M}$, and $\Delta_{\mathcal{M}}$ denotes the Laplace-Beltrami operator defined by $\Delta f = \text{div}(\nabla f)$. $\exp_x:T_x\mathcal{M}\rightarrow\mathcal{M}$ and $\exp_x^{-1}:\mathcal{M}\rightarrow T_x\mathcal{M}$ denotes the Riemannian exponential and logarithm map, respectively. Lastly, $\mathrm{d}\text{vol}_x$ denotes the volume form on the manifold, and $\int\! f(x)\mathrm{d}\text{vol}_x$ is the integration of function $f$ on the manifold.

\paragraph{Diffusion Process on Riemannian Manifold}
Brownian motion on a Riemannian Manifold $\mathcal{M}$ is a diffusion process generated by $\Delta_{\mathcal{M}}/2$~\citep{hsu2002stochastic} which is a generalization of the Euclidean Brownian motion. The transition distribution of the Brownian motion corresponds to the heat kernel, i.e. the solution to the heat equation, which coincides with the Gaussian distribution when $\mathcal{M}$ is a Euclidean space. One can construct a diffusion process that converges to a stationary distribution described by the Langevin dynamics: 
\begin{align}
    \mathrm{d}\bm{X}_t = -\frac{1}{2}\nabla_{\bm{X}_t}U(\bm{X}_t) \mathrm{d}t + \mathrm{d}\mathbf{B}^{\mathcal{M}}_t,
\label{eq:noising_process}
\end{align}
where $\mathbf{B}^{\mathcal{M}}_t$ denotes the Brownian motion defined on $\mathcal{M}$, such that the terminal distribution satisfies $\mathrm{d}p(x)/\mathrm{d}\text{vol}_x\propto e^{-U(x)}$~\citep{durmus2016high} for a potential function $U$ which we describe in detail in Appendix~\ref{app:derivations:bb}.
A diffusion process on the manifold can be simulated using the Geodesic Random Walk~\citep{jorgensen1975central, debortoli2022rsgm} which corresponds to taking a small step on the tangent space in the direction of the drift.
\section{Riemannian Diffusion Mixture}
We now present Riemannian Diffusion Mixture, a new framework for learning a generative diffusion process on Riemannian manifolds using a mixture of bridge processes.

\subsection{Bridge Processes on Manifold}
The first step of constructing the generative process is designing a diffusion process conditioned to fixed endpoints, i.e. the bridge process. In contrast to the Euclidean space which is equipped with simple families of bridge processes derived from the Brownian motion or the Ornstein-Uhlenbeck process~\citep{peluchetti2023mixture,jo2023graph}, designing a bridge process on general manifolds is challenging since the transition density of the Brownian motion is intractable in general. 

To achieve a simple bridge process that can be used for building a generative model, we start with the Brownian bridge on the manifold $\mathcal{M}$ with fixed endpoints, modeled by the following SDE (see Appendix~\ref{app:derivations:bb} for details of the Brownian bridge):
\begin{align}
    \mathrm{d}\bm{X}_t = \nabla_{\!\bm{X}_t}\! \log p_{\mathcal{M}}(\bm{X}_t, z, T\!-\!t)\mathrm{d}t + \mathrm{d}\mathbf{B}^{\mathcal{M}}_t ,
\label{eq:brownian_bridge}
\end{align}
for $\bm{X}_0=z_0$ where $p_{\mathcal{M}}$ denotes the heat kernel on $\mathcal{M}$ and $z$ denotes the fixed endpoint. We cannot directly use this Brownian bridge process as the heat kernel is known in very limited cases and even on a simple manifold such as a sphere, the heat kernel is represented as an infinite sum~\citep{tulovsky2001formula}. Thereby we explore a new family of bridge processes that do not require the heat kernel.

\begin{figure}[!t]
    \centering
    \centerline{\includegraphics[width=0.9\linewidth]{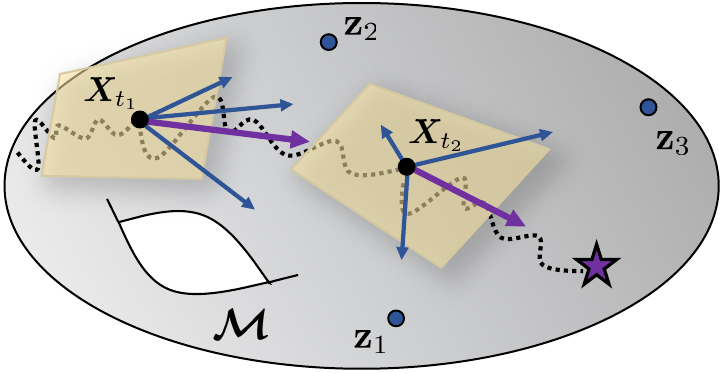}}
    \vspace{-0.1in}
    \caption{We construct a generative process on general manifolds as a mixture of bridge processes (Eq.~\eqref{eq:mixture_process}). The drift of the mixture process (\textcolor{figpurple}{\textbf{purple vector}}) corresponds to the weighted mean of the tangent vectors pointing to the directions of the endpoints (\textcolor{figblue}{\textbf{blue vector}}), guiding the diffusion process (black dotted) to the data distribution.
    }
    \label{fig:concept}
    \vspace{-0.1in}
\end{figure}

Intuitively, a diffusion process that takes each step in the direction of the endpoint should carry the process toward the desired endpoint regardless of the prior distribution. The natural choice for this direction would be following the shortest path between the current state and the endpoint on the manifold, which corresponds to the inverse of the exponential map\footnote{Here we assume that the endpoint is not in the cut locus of the current state for the inverse to be well-defined.}, i.e., the logarithm map. The logarithm map provides a simple approach to represent a tangent vector that heads toward the desired endpoint, as illustrated in Figure~\ref{fig:concept} by the blue vectors pointing to the endpoints.

From this observation, we introduce a simple family of bridge processes on manifolds derived from the logarithm map, namely the \emph{Logarithm Bridge Process} $\mathbb{Q}^{x,z}_{log}$:
\begin{align}
    \mathrm{d}\bm{X}_t = \frac{\sigma_t^2}{\tau_T-\tau_t} \exp^{\scalebox{0.75}[1.0]{-}1}_{\!\bm{X}_t}\!(z) \mathrm{d}t + \sigma_t\mathrm{d}\mathbf{B}^{\mathcal{M}}_t
    \label{eq:logarithm_bridge}
\end{align}
for $\bm{X}_0\!=\!x$, where $z$ is the fixed endpoint, $\exp^{\scalebox{0.75}[1.0]{-}1}$ is the logarithm map, $\sigma_t$ is the time-dependent noise schedule that uniquely determines the process, and $\tau(t) \coloneqq \int^t_0 \! \sigma_s^2\mathrm{d}s$ denotes the rescaled time with respect to $\sigma_t$. 
A key observation is that the logarithm map $\exp^{\scalebox{0.75}[1.0]{-}1}$ represents the direction of the shortest path between the current state $\bm{X}_t$ and the endpoint $z$. As the magnitude of the drift increases to infinity with a rate $\sigma^2_t/(\tau_T-\tau_t)$ as $t\rightarrow T$, the process is forced to converge to $z$ by the direction of the drift.

By leveraging the short-time asymptotic behavior of the Brownian motion and the Girsanov theorem, we theoretically derive in Appendix~\ref{app:derivations:lbp} that the Logarithm bridge exhibits a similar convergence behavior as the Brownian bridge, regardless of the prior distribution $\Gamma$. 
In particular, when $\mathcal{M}$ is a Euclidean space $\mathbb{R}^d$, the logarithm bridge process reduces to the well-known Euclidean Brownian bridge process. But for general manifolds, our Logarithm bridge process differs from the Brownian bridge process of Eq.~\eqref{eq:brownian_bridge} due to the difference in the drifts.

Although the Logarithm bridge provides a simple solution for constructing the generative process on manifolds, the logarithm map of general manifolds may not be given in closed form and could be costly to compute on the fly. We can bypass the difficulty by taking a different perspective for defining the direction toward the endpoint on the manifold. Specifically, inspired by \citet{chen2024rfm}, we consider a path on the manifold that minimizes the spectral distance $d_w(\cdot,\cdot)$, which is defined by the eigenvalues $\lambda_i$ and eigenfunctions $\phi_i$ of the Laplace-Beltrami operator $\Delta_{\mathcal{M}}$:
\begin{align}
    d_w(x,y)^2 = \sum^{\infty}_{i=1} w(\lambda_i)\big( \phi_i(x) - \phi_i(y) \big)^2,
\label{eq:spec_distance}
\end{align}
where $w$ is a monotonically decreasing function.  
From the fact that $\nabla d_w(\cdot,z)^2$ describes the tangent vector at the current state with the direction that minimizes the spectral distance to the endpoint $z$, we introduce a new family of bridge processes, namely \emph{Spectral Bridge Process} $\mathbb{Q}^{x, z}_{spec}$:
\begin{align}
    \mathrm{d}\bm{X}_t = -\frac{1}{2}\frac{\sigma_t^2}{\tau_T-\tau_t} \frac{\nabla_{\!\!\bm{X}_t}\! d_w(\bm{X}_t, z)^2}{\left\| \nabla_{\!\!\bm{X}_t}\! d_w(\bm{X}_t, z) \right\|^2_{\mathcal{M}}} \mathrm{d}t + \sigma_t\mathbf{B}^{\mathcal{M}}_t 
\label{eq:spec_bridge}
\end{align}
for $\bm{X}_0\!=\!x$ where the norm of the gradient $\nabla d_w$ normalizes the magnitude of the drift.
The Spectral bridge process in Eq.~\eqref{eq:spec_bridge} is designed so that substituting the spectral distance $d_w$ with the geodesic distance $d_g$ results in the Logarithm bridge process in Eq.~\eqref{eq:logarithm_bridge}.
Note that the eigenvalues and the eigenfunctions are computed only once, in advance of training our generative model, and do not require computing eigenfunctions during training.

%%%%%%%%%%%%%%%%%%%%%%%%%%%%%%%%%%%%%%%
%%%%%%%%%%%%%%%%%%%%%%%%%%%%%%%%%%%%%%%

Other choices of bridge processes on manifolds are possible, such as semi-classical Brownian bridge~\citep{elworthy1981classical} or Fermi bridge~\citep{thompson2015submanifold} which we describe in Appendix~\ref{app:derivations:general_bridge}. 
Yet we focus on the Logarithm bridge and Spectral bridge due to their simplicity and practicality for real-world problems where the most relevant manifolds either have known geodesics or eigenfunctions of the Laplace-Beltrami operator, for example, $SE(3)^N$ for protein modeling~\citep{jumper2021highly}, $SU(N)$ for high energy physics~\citep{boyda2021sampling}, and product of tori for molecular conformed generation~\citep{jing2022torsion}.
We further note that one can consider using the Brownian bridge (Eq.~\eqref{eq:brownian_bridge}) with appropriate estimation of the heat kernel, e.g., Varadhan approximation~\citep{debortoli2022rsgm} or eigenfunction expansion on maximum torus~\citep{lou2023scaling}.

\subsection{Generative Process on Riemannian Manifold}

\paragraph{Riemannian Diffusion Mixture}
Having the bridge processes at hand, we now build a diffusion process on manifolds that transports a prior distribution $\Gamma$ to the data distribution $\Pi$. 
Extending the diffusion mixture framework~\citep{peluchetti2021mixture, liu2023bridge} to the Riemannian setting, we construct a generative process on the manifold $\mathcal{M}$ by mixing a collection of bridge processes on $\mathcal{M}$, denoted as $\{\mathbb{Q}^{x,z}\!:\!x\!\sim\!\Gamma, z\!\sim\!\Pi\}$.

We derive a diffusion process that admits a marginal density $p_t$ which is equal to the mixture of marginal densities $p^{x,z}_t$ of $\mathbb{Q}^{x,z}$, that is modeled by the following SDE (we provide a formal definition and a detailed derivation in Appendix~\ref{app:derivations:mix}):
\begin{align}
    \mathrm{d}\bm{X}_t 
    = \left[ \int 
    \!\! \eta^{z}\!(\bm{X}_t, t) \frac{p^{z}_t(\bm{X}_t)}{p_t(\bm{X}_t)} \Pi(\mathrm{d}\text{vol}_z) \right] 
    \!\mathrm{d}t + \sigma_t\mathrm{d}\mathbf{B}^{\mathcal{M}}_t
\label{eq:mixture_process}
\end{align}
for $X_0\!\sim\!\Gamma$ where $\eta^{z}$ denotes the drift of $\mathbb{Q}^{x,z}$, $p^{z}_t(\cdot)\!=\!\int\! p^{x,z}_t(\cdot)\Gamma(\mathrm{d}\text{vol}_x)$, and $p_t(\cdot)\!=\!\int\! p^{z}_t(\cdot)\Pi(\mathrm{d}\text{vol}_z)$.

From the geometrical viewpoint, the drift of the mixture process (Figure~\ref{fig:concept} \textcolor{figpurple}{purple}) corresponds to the weighted mean of tangent vectors (Figure~\ref{fig:concept} \textcolor{figblue}{blue}) heading in the direction of the points in the data distribution.
Therefore, simulating the mixture process can be interpreted as taking a small step in the direction of the most likely endpoint of the process. 
From this perspective, we can derive an explicit prediction from the mixture process, in particular, from the mixture of Logarithm bridges by projecting the drift onto the manifold along the geodesic using the exponential map as follows:
\begin{align}
    \hat{\bm{X}}_t \coloneqq \exp_{\bm{X}_t} \left( \int \exp^{\scalebox{0.75}[1.0]{-}1}_{\!\bm{X}_t}\!(z) \frac{p^{z}_t(\bm{X}_t)}{p_t(\bm{X}_t)}\, \Pi(\mathrm{d}\text{vol}_z) \right) ,
\label{eq:prediction}
\end{align}
which corresponds to the most probable endpoint of the mixture process given the current state. This differentiates our method from the previous diffusion models that do not admit straightforward predictions for their denoising process on non-Euclidean manifolds.

Notably, by the construction of the mixture process, its terminal distribution is guaranteed to be equal to the data distribution $\Pi$ regardless of the initial distribution $\Gamma$, and thereby our framework can be trivially applied to an arbitrary initial distribution. This is not the case for previous diffusion models~\citep{debortoli2022rsgm, huang2022rdm} which require careful design of the potential $U(\cdot)$ in the noising process (Eq.~\eqref{eq:noising_process}), since they rely on the denoising diffusion framework, in contrast to our bridge mixture construction.
Our framework yields freedom for the choice of the noise schedule $\sigma_t$ in Eq.~\eqref{eq:mixture_process} where $\sigma_t$ need not be decreasing or be large for small $t$.

When the mixture consists of the Logarithm bridges or the Spectral bridges, we refer to the mixture processes as the \emph{Logarithm Bridge Mixture} (LogBM) and \emph{Spectral Bridge Mixture} (SpecBM), respectively. Note that our LogBM generalizes previous diffusion mixture framework~\citep{peluchetti2021mixture, liu2023bridge} since the Logarithm bridge recovers the Brownian bridge for the Euclidean space.

\paragraph{Probability Flow ODE}
For a mixture process $\mathbb{Q}_f$, there exists a deterministic process that admits the same marginal densities, i.e., the probability flow~\citep{maoutsa2020interacting, song2021sde}. 
The time-reversed process $\mathbb{Q}_b$ of $\mathbb{Q}_f$ is also a mixture process built from the collection of time-reversed bridge processes, which can be derived from $\mathbb{Q}_f$ in terms of the score function (Eq.~\eqref{eq:reversed_mixture_process_score}). As a result, the probability flow associated with $\mathbb{Q}_f$ satisfies the following ODE (see Appendix~\ref{app:derivations:pode} for detailed derivation):
\begin{align}
    \frac{\mathrm{d}}{\mathrm{d}t} \bm{Y}_t = \frac{1}{2} \Big( \eta_f(\bm{Y}_t,t) - \eta_b(\bm{Y}_t,T\!-\!t) \Big) \;,\;\; \bm{Y}_0\sim \Gamma ,
\label{eq:probability_flow}
\end{align}
where $\eta_f$ and $\eta_b$ denote the drift of $\mathbb{Q}_f$ and $\mathbb{Q}_b$, respectively, and the likelihood of the probability flow as follows:
\begin{align}
    &\log p_T(\bm{Y}_T) - \log p_0(\bm{Y}_0) \notag \\
    &= \frac{1}{2} \int^{T}_0 \text{div}\Big( \eta_f(\bm{Y}_t,t) - \eta_b(\bm{Y}_{t},T\!-\!t) \Big) \mathrm{d}t.
\label{eq:pode_likelihood}
\end{align}
We further discuss in Appendix~\ref{app:derivations:pode} that the probability flow derived from our mixture process is different from the continuous flows used in previous CNF models.

\subsection{Two-way Bridge Matching \label{sec:method:generative_model}}
Now we show how to train a generative model that approximates the mixture process in Eq.~\eqref{eq:mixture_process}. We parameterize the drifts of the mixture process and its time-reversed process with neural networks, i.e., $\bm{s}^{\theta}_f(z,t) \!\approx\! \eta_f(z,t) $ and $\bm{s}^{\phi}_b(z,t) \!\approx\! \eta_b(z,t)$. However, the drifts of the mixture processes cannot be directly approximated since we do not have access to the integration in Eq.~\eqref{eq:mixture_process}. In what follows, we derive a simple and efficient training objective that is applicable to general manifolds without computing the Riemannian divergence.

For a diffusion process $\mathbb{Q} \!:\! \mathrm{d}\bm{Z}_t \!=\! \eta(\bm{Z}_t,t)\mathrm{d}t + \nu_t \mathrm{d}B^{\mathcal{M}}_t$ and its parameterized process $\mathbb{P}^{\psi} \!:\! \mathrm{d}\bm{Z}_t = s^{\psi}\!(\bm{Z}_t,t)\mathrm{d}t + \nu_t\mathrm{d}B^{\mathcal{M}}_t$, 
the KL divergence between two processes can be obtained from the Girsanov theorem as follows (we provide detailed derivation in Appendix~\ref{app:derivations:bridge_matching}):
\begin{align}
    &D_{KL}(\mathbb{Q}_T \| \mathbb{P}^{\psi}_T) \leq
    D_{KL}(\mathbb{Q} \| \mathbb{P}^{\psi}) \label{eq:objective_naive} \\
    &= \mathbb{E}_{\substack{z\sim \mathbb{Q}_T \\ \bm{Z}\sim\mathbb{Q}^z}} \!\! \left[ \frac{1}{2}\! \int^T_0 \!\! \left\| \nu_t^{-1} \Big(\bm{s}^{\psi}(\bm{Z}_t,t) - \eta^{z}(\bm{Z}_t,t) \Big) \right\|^2_{\mathcal{M}} \!\! \mathrm{d}t\right] + C \notag
\end{align}
where $\mathbb{Q}_T$ and $\mathbb{P}^{\psi}_T$ denotes the terminal distributions of $\mathbb{Q}$ and $\mathbb{P}^{\psi}$ respectively, $\mathbb{Q}^z$ and $\eta^{z}$ denotes the process $\mathbb{Q}(\cdot|\bm{Z}_T\!=\!z)$ and its drift, and $C$ is a constant. 

Although we can directly use Eq.~\eqref{eq:objective_naive} to train $s^{\theta}_f$ and $s^{\phi}_b$, it is computationally expensive as the samples $\bm{Z}_t$ should be obtained through a simulation of bridge processes. This is because the transition density of the Brownian motion is not accessible for general manifolds. Especially, simulating the bridge process requires a large number of discretized steps due to the exploding magnitude of the drift of the bridge process near the terminal time, i.e., magnitude of the drift explodes with a rate $\sigma^2_t/(\tau_T-\tau_t)$ as $t\rightarrow T$.

We address this issue by proposing an efficient training scheme, which we refer to as the \emph{two-way bridge matching}. The main idea is to exploit the fact that (1) the simulation of the bridge process can be performed from both forward and backward directions, which can bypass the explosion of drift that allows larger step size, and (2) $\bm{Z}_t$ can be obtained from a single bridge process with fixed endpoints instead of simulating two different bridge processes, reducing the computational cost in half.
Altogether, the two-way bridge matching is formalized as follows:
\begin{align}
    &\mathbb{E}_{\substack{(x,y)\sim (\Pi,\Gamma), \\ t \sim [0,T]}}
    \mathbb{E}_{\bm{Z}_t \sim \mathbb{Q}^{x,y}} \;
    \sigma^{-2}_t \bigg[ 
        F^{\theta,x}_f(\bm{Z}_t,t) + F^{\phi,y}_b(\bm{Z}_t, t)
    \bigg] \notag \\
    & F^{\theta,x}_f(\bm{Z}_t, t) = \Big\| \bm{s}^{\theta}_f(\bm{Z}_t,t) - \eta^x_f(\bm{Z}_t,t) \Big\|^2_{\mathcal{M}}, \label{eq:twoway_bridge_matching} \\
    & F^{\phi,y}_b(\bm{Z}_t, t) = \Big\| \bm{s}^{\phi}_b(\bm{Z}_t,T\!\!-\!t) - \eta^y_b(\bm{Z}_t,T\!\!-\!t) \Big\|^2_{\mathcal{M}}, \notag
\end{align}
where $\mathbb{Q}^{x,y}$ denotes the bridge process with fixed starting point $x$ and endpoint $y$, and $\bm{Z}_t$ is obtained in a two-way approach, i.e., simulating $\mathbb{Q}^{x,y}$ from time $0$ to $t$ if $t<T/2$, and otherwise simulating from time $T$ to $t$.
Notably, from the result of Eq.~\eqref{eq:objective_naive}, minimizing Eq.~\eqref{eq:twoway_bridge_matching} guarantees to minimize the KL divergence between data distribution and terminal distribution of our parameterized mixture process.

However, we empirically observe that using Eq.~\eqref{eq:twoway_bridge_matching} introduces high variance during training due to the coefficient $\sigma_t^{-2}$. 
Therefore, we present an equivalent objective that enables stable training by leveraging importance sampling, where we use a proposal distribution $q(t)\propto\sigma_t^{-2}$ to adjust the weighting as follows:
\begin{align}
\hspace*{-1mm}
    \mathbb{E}_{\substack{(x,y)\sim (\Pi,\Gamma), \\ t \sim q}}
    \mathbb{E}_{\bm{Z}_t \sim \mathbb{Q}^{x,y}}
    \bigg[
    F^{\theta,x}_f(\bm{Z}_t,t) + F^{\phi,y}_b(\bm{Z}_t, t)
    \bigg],
\label{eq:importance_sampling}
\hspace*{-1mm}
\end{align}
which we refer to as the \emph{time-scaled} two-way bridge matching. During training with Eq.~\eqref{eq:importance_sampling}, we first sample $t\sim q$ and $(x,y)\sim (\Pi,\Gamma)$, then simulate $\bm{Z}_t\sim\mathbb{Q}^{x,y}$ using the two-way approach, where different triplets $(t,x,y)$ are sampled to compute the expectation.
We summarize the training process of our two-way bridge matching in Algorithm~\ref{alg:training}.

%%%%%%%%%%%%%%%%%%%%%%%%%%%%%%
\begin{figure}[t!]
\vspace{-0.1in}
\centering
\begin{algorithm}[H]
    \caption{Two-way bridge matching}\label{alg:training}
        \textbf{Input:} Training set $\mathcal{D}$, prior distribution $\Gamma$, trained neural networks $\bm{s}^{\theta}_f$ and $\bm{s}^{\phi}_b$, terminal time T, number of in-training simulation step $N$ \\
        \textbf{For each epoch:} \phantom{a}
    \begin{algorithmic}[1]
        \STATE Sample $x\sim \mathcal{D}$, $y\sim\Gamma$, and $t\sim q $
        \STATE $(z_0, z_f, \eta)\leftarrow (y, x, \eta_f)$\; {if}\; $t\!<\!T/2$\; {else}\; $(x, y, \eta_b)$
        % \STATE $(z, s, \mathrm{d}t) \!\leftarrow\! (y, 0, t/N)$ {if} $t\!<\!T/2$ {else} $(x, T, -t/N)$ 
        \STATE $\mathrm{d}t \leftarrow t/N$
        \STATE $z \leftarrow z_0$, $s\leftarrow 0$, $\mathrm{d}t \leftarrow t/N$ 
        \FOR[In-training simulation of $\bm{Z}_t$]{$n=1$ \textbf{to} $N$}
            \STATE $W \sim \mathcal{N}(0,\text{Id})$ 
            \COMMENT{Random normal in $T_z\mathcal{M}$}
            \STATE $v \leftarrow \eta^{z_f}(z,s)\mathrm{d}t + \sigma_t W \sqrt{\mathrm{d}t}$
            % \COMMENT{Euler-Maruyama step on tangent space}
            \STATE $z\leftarrow \exp_{z} v$
            \COMMENT{Geodesic step in direction of $v$}
            \STATE $s \leftarrow s + \mathrm{d}t$
        \ENDFOR
        \STATE $\mathcal{L}_{\theta,\phi} \leftarrow F^{\theta,x}_f(z,t) + F^{\phi,y}_b(z,t)$
        \COMMENT{Eq.~\eqref{eq:importance_sampling}}
        \STATE Update $\theta$, $\phi$ using $\mathcal{L}_{\theta,\phi}$
    \end{algorithmic}
\end{algorithm}
\vspace{-0.3in}
\end{figure}
%%%%%%%%%%%%%%%%%%%%%%%%%%%%%%

In particular, our two-way bridge matching consists of a simple regression on the drifts of bridge processes, i.e., $F^{\theta, x}_f$ and $F^{\phi,y}_b$, without computation of divergence or any approximation. Thereby, on manifolds for modeling real-world problems, our framework can scale to high dimensions, which previous generative models cannot scale to.

We note that our time-scaled two-way bridge matching differs from the Flow Matching objective which regresses the conditional vector field over uniformly distributed time. We experimentally validate the importance of the time distribution $q$ in Section~\ref{sec:exp:abl}, where using a uniform time distribution results in a significant drop in performance compared to using $t\!\sim\!q$. 
This is because Eq.~\eqref{eq:importance_sampling} guarantees to maximize the likelihood of our generative model, whereas it is not true for a simple regression over uniformly distributed time.

We empirically validate that our two-way approach can obtain accurate trajectories with significantly reduced simulation steps compared to the one-way simulation, resulting in up to $\times$34.9 speed up for training. Furthermore, we show in Section~\ref{sec:exp} that the in-training simulation is not a significant overhead during training since the two-way approach greatly reduces the number of simulation steps without sacrificing the accuracy, which is significantly faster than the implicit score matching used for previous diffusion models.

\paragraph{Connection with Riemannian Flow Matching}
Especially, in the case when the noise schedule is set to be very small, i.e., $\sigma_t\rightarrow 0$, we can recover the deterministic flow of Riemannian Flow Matching (RFM)~\citep{chen2024rfm} from our mixture process, where the bridge processes with $\sigma_t\rightarrow 0$ correspond to the conditional vector fields of Flow Matching. Thereby RFM can be considered a special case of our framework when the randomness is removed.

However, stochasticity is crucial for learning the density on manifolds with non-trivial curvature. While obtaining a trajectory of the probability path for RFM during training requires a large number of simulation steps, we can obtain trajectories from the mixture process with only a few simulation steps thanks to its stochastic nature, which we empirically show in Section~\ref{sec:exp:abl}. 
The existence of stochasticity dramatically reduces the number of in-training simulation steps compared to RFM, achieving $\times$12.8 speed up in training without sacrificing the performance. We demonstrate in Figures~\ref{fig:mesh_vis} and \ref{fig:mesh_vis_abl} that our method is able to model complex distribution on the manifold with only a few in-training simulation steps, whereas RFM completely fails in such a setting.
Furthermore, we can leverage the Girsanov theorem to derive that our training objective in Eq.~\eqref{eq:importance_sampling} is guaranteed to minimize the KL divergence between the data distribution and the terminal distribution of our parameterized process, which does not apply to RFM as it is based on a deterministic process.

\paragraph{Sampling}
Generating samples with the Riemannian diffusion mixture can be achieved in two different ways: (1) simulating the approximation of the mixture process (Eq.~\eqref{eq:mixture_process}) and (2) simulating the probability flow (Eq.~\eqref{eq:probability_flow}).
First, the mixture process can be approximated by using the drift estimation $\bm{s}^{\theta}_f$ as follows:
\begin{align}
    \mathrm{d}\bm{X}_t = \bm{s}^{\theta}_f(\bm{X}_t,t) \mathrm{d}t + \sigma_t\mathrm{d}\mathbf{B}^{\mathcal{M}}_t.
\end{align}
which can be simulated using the Geodesic Random Walk~\citep{jorgensen1975central,debortoli2022rsgm}.
Alternatively, the probability flow can be modeled by the ODE using the drift estimations $\bm{s}^{\theta}_f$ and $\bm{s}^{\phi}_b$ as follows:
\begin{align}
    \frac{\mathrm{d}}{\mathrm{d}t} \bm{X}_t = \frac{1}{2}\left( \bm{s}^{\theta}_f(\bm{X}_t,t) -  \bm{s}^{\phi}_b(\bm{X}_t,T\!-\!t) \right) ,
\end{align}
which can be solved using integrators on Riemannian manifolds~\citep{hairer2011solving}.
\section{Related Work \label{sec:related_work}}

\paragraph{Euclidean Diffusion Models}
Diffusion models~\citep{song2019smld, ho2020ddpm, song2021sde} model the generative process via the denoising diffusion process derived from the time-reversal of the noising process. Recent works~\citep{peluchetti2021mixture, liu2023bridge, peluchetti2023mixture} introduce an alternative approach for modeling the generative process without using the time-reversal, namely diffusion mixture, by building bridge processes between the initial and the terminal distributions. However, these methods are limited to Euclidean space and sub-optimal for modeling the data living on manifolds, for example, sphere for climate data and tori for biological data such as proteins. Our work shows how to extend the diffusion mixture framework to manifolds that generalize the Euclidean case.

\paragraph{Generative Models on Riemannian Manifolds \label{sec:related_work:manifold_model}}
Previous generative models~\citep{gemici2016normalizing, rezende2020nftori, bose2020hyperbolic} relied on projecting a Euclidean space to manifolds which is problematic since such mapping cannot be bijective, resulting in numerical instabilities. Recent works address this problem by constructing a mapping on the manifold that describes the transport from a prior distribution to the data distribution, namely the diffusion models and the CNF models.

The seminal work of \citet{debortoli2022rsgm} extends the score-based model to the manifold, while \citet{huang2022rdm} introduces a variational framework for diffusion models on manifolds. However, both rely on score matching that either needs to be approximated or scales poorly to higher dimensions. Specifically, denoising score matching requires the conditional score function to be approximated which obstructs exact training. Further, implicit score matching requires the computation of the Riemannian divergence which scales poorly to high dimensions, and using the Hutchinson estimator~\citep{hutchinson1989stochastic} introduces high variance in training.
In contrast, our framework provides efficient and scalable training that does not require divergence and does not rely on approximations of the heat kernel. Since the construction of the mixture process guarantees convergence to the data distribution regardless of the prior distribution, our method can be readily extended for arbitrary prior distribution. We provide discussion on Diffusion Schr\"odinger Bridge~\citep{thornton2022rdsb}, improvement of Riemannian diffusion models~\citep{lou2023scaling}, and recent works focusing on specific geometries in Appendix~\ref{app:derivations:comparison}.

%%%%%%%%%%%%%%%%%%%%%%%%%%%%%%%%%%%%%%%%%
\begin{table*}[t]
% \vspace{-0.075in}
\caption{
\textbf{Test NLL results on earth and climate science datasets}. We report the mean of 5 different runs with different data splits. Best performance and its comparable results ($p>0.05$) from the t-test are highlighted.}
\label{tab:earth}
\vspace{-0.075in}
\centering
    \resizebox{\textwidth}{!}{
    \renewcommand{\arraystretch}{0.9}
    \renewcommand{\tabcolsep}{14pt}
\begin{tabular}{l c c c c}
\toprule
     & Volcano & Earthquake & Flood & Fire \\
     Dataset size & 827 & 6120 & 4875 & 12809 \\
\midrule
    RCNF~\citep{mathieu2020rcnf} & -6.05~\scriptsize{$\pm$ 0.61} & 0.14~\scriptsize{$\pm$ 0.23} & 1.11~\scriptsize{$\pm$ 0.19} & -0.80~\scriptsize{$\pm$ 0.54} \\
    Moser Flow~\citep{rozen2021moser} & -4.21~\scriptsize{$\pm$ 0.17} & -0.16~\scriptsize{$\pm$ 0.06} & 0.57~\scriptsize{$\pm$ 0.10} & -1.28~\scriptsize{$\pm$ 0.05} \\
    CNFM~\citep{ben-hamu2022cnfm} & -2.38~\scriptsize{$\pm$ 0.17} & \textbf{-0.38}~\scriptsize{$\pm$ 0.01} & \textbf{0.25}~\scriptsize{$\pm$ 0.02} & -1.40~\scriptsize{$\pm$ 0.02} \\
    RFM~\citep{chen2024rfm} & -7.93~\scriptsize{$\pm$ 1.67} & -0.28~\scriptsize{$\pm$ 0.08} & 0.42~\scriptsize{$\pm$ 0.05} & -1.86~\scriptsize{$\pm$ 0.11} \\
\midrule
    StereoSGM~\citep{debortoli2022rsgm} & -3.80~\scriptsize{$\pm$ 0.27} & -0.19~\scriptsize{$\pm$ 0.05} & 0.59~\scriptsize{$\pm$ 0.07} & -1.28~\scriptsize{$\pm$ 0.12} \\
    RSGM~\citep{debortoli2022rsgm} & -4.92~\scriptsize{$\pm$ 0.25} & -0.19~\scriptsize{$\pm$ 0.07} & 0.45~\scriptsize{$\pm$ 0.17} & -1.33~\scriptsize{$\pm$ 0.06} \\
    RDM~\citep{huang2022rdm} & -6.61~\scriptsize{$\pm$ 0.96} & \textbf{-0.40}~\scriptsize{$\pm$ 0.05} & 0.43~\scriptsize{$\pm$ 0.07} & -1.38~\scriptsize{$\pm$ 0.05} \\
    RSGM-improved~\citep{lou2023scaling} & -4.69~\scriptsize{$\pm$ 0.29} & -0.27~\scriptsize{$\pm$ 0.05} & 0.44~\scriptsize{$\pm$ 0.03} & -1.51~\scriptsize{$\pm$ 0.13} \\
\midrule
    Ours (LogBM) & \textbf{-9.52}~\scriptsize{$\pm$ 0.87} & -0.30~\scriptsize{$\pm$ 0.06} & 0.42~\scriptsize{$\pm$ 0.08} & \textbf{-2.47}~\scriptsize{$\pm$ 0.11} \\
\bottomrule
\end{tabular}}
\vspace{-0.05in}
\end{table*}
\begin{figure*}[!t]
\centering
\vspace{-0.1in}
\begin{minipage}{0.63\linewidth}
\resizebox{\textwidth}{!}{
    \renewcommand{\arraystretch}{1.1}
    \renewcommand{\tabcolsep}{5pt}
\begin{tabular}{l c c c}
\toprule
     & Glycine (2D) & Proline (2D) & RNA (7D) \\
     Dataset size & 13283 & 7634 & 9478 \\
\midrule
    MoPS~\citep{de2020power} & 2.08~\scriptsize{$\pm$ 0.009} & 0.27~\scriptsize{$\pm$ 0.008} & 4.08~\scriptsize{$\pm$ 0.368} \\
    RDM~\citep{huang2022rdm} & 1.97~\scriptsize{$\pm$ 0.012} & \textbf{0.12}~\scriptsize{$\pm$ 0.011} & -3.70~\scriptsize{$\pm$ 0.592} \\
    RFM~\citep{chen2024rfm} & \textbf{1.90}~\scriptsize{$\pm$ 0.055} & 0.15~\scriptsize{$\pm$ 0.027} & \textbf{-5.20}~\scriptsize{$\pm$ 0.067} \\
\midrule
    Ours (LogBM) & \textbf{1.89}~\scriptsize{$\pm$ 0.056} & \textbf{0.14}~\scriptsize{$\pm$ 0.027} & \textbf{-5.27}~\scriptsize{$\pm$ 0.090} \\
\bottomrule
\end{tabular}}
\end{minipage}
\hfill
\begin{minipage}{0.34\linewidth}
    \includegraphics[width=1\linewidth]{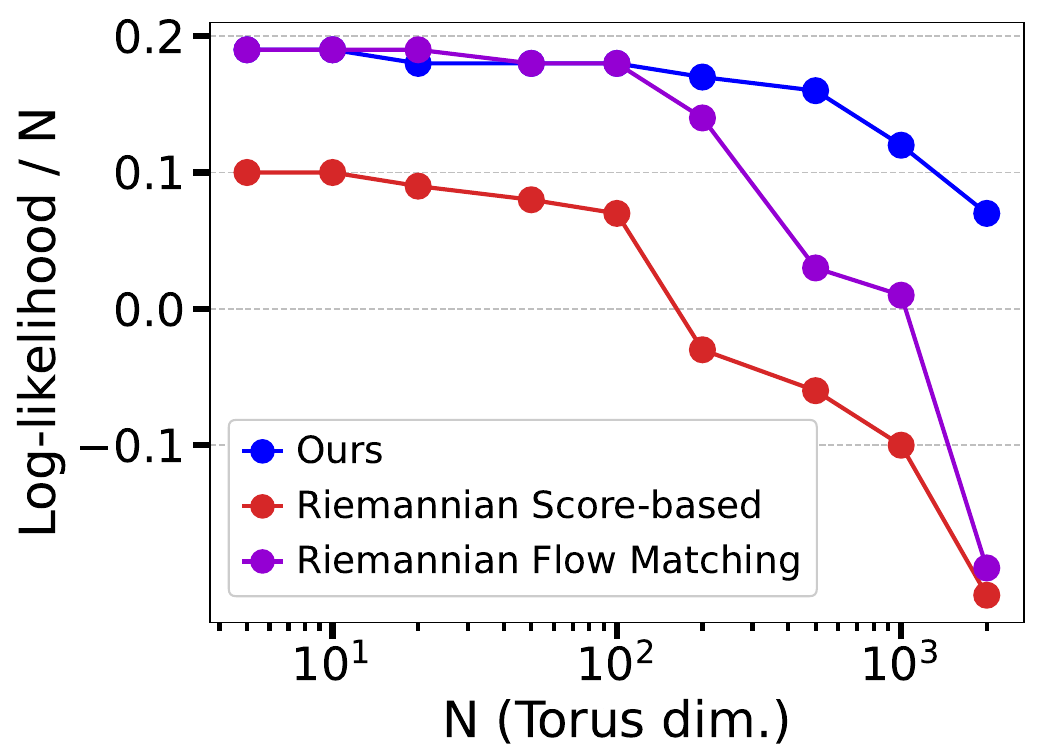}
\end{minipage}
\vspace{-0.125in}
\caption{\textbf{(Left) Test NLL results on protein datasets}. Best performance and its comparable results ($p>0.05$) from the t-test are highlighted in bold. 
\textbf{(Right) Comparison on high-dimensional tori}. We compare the log-likelihood in bits against RSGM and RFM where the results are obtained by running the open-source codes.}
\label{fig:torus}
\vspace{-0.125in}
\end{figure*}
%%%%%%%%%%%%%%%%%%%%%%%%%%%%%%%%%%%%%%%%%

On the other hand, CNF models build a continuous-time flow~\citep{chen2018node, grathwohl2019ffjord} on the manifold by parameterizing the vector field. However, previous CNF models~\citep{lou2020neural, mathieu2020rcnf, falorsi2020neural} rely on simulation-based maximum likelihood training which is computationally expensive. Recent works~\citep{rozen2021moser, ben-hamu2022cnfm} introduce simulation-free training methods on simple geometries, but they scale poorly to high-dimension due to the computation of the divergence and further cannot be adapted to non-simple geometries. \citet{chen2024rfm} extends the Flow Matching framework~\citep{lipman2023flowmatching} to manifolds which learns the probability path by regressing the conditional vector fields. 
Instead of the deterministic flow, our work constructs a diffusion-based generative process for which stochasticity is crucial for learning on general geometries, as it achieves superior performance with greatly reduced in-training simulation steps.
\section{Experiments \label{sec:exp}}

We experimentally validate our method on diverse datasets including real-world benchmarks as well as synthetic distributions. We follow the experimental settings of previous works~\citep{debortoli2022rsgm, chen2024rfm} where we provide the details of the training setup in Appendix~\ref{app:exp}. We compare our method against generative models on manifolds: \textbf{RCNF}~\citep{mathieu2020rcnf}, \textbf{Moser Flow}~\citep{rozen2021moser}, \textbf{CNFM}~\citep{ben-hamu2022cnfm} and \textbf{RFM}~\citep{chen2024rfm} are CNF models, \textbf{StereoSGM}~\citep{debortoli2022rsgm} is a Euclidean score-based model using stereographic projection, \textbf{RSGM}~\citep{debortoli2022rsgm} is a Riemannian score-based model, \textbf{RDM}~\citep{huang2022rdm} is a Riemannian diffusion model based on a variational framework, and \textbf{RSGM-improved}~\citep{lou2023scaling} uses improved heat kernel estimator for RSGM.

%%%%%%%%%%%%%%%%%%%%%%%%%%%%%%%%%%%%%%%%
\begin{figure*}[!t]
\centering
\caption{\textbf{Visualization of the generated samples and the learned density} of our method and RFM on the mesh datasets. Blue dots represent the generated samples and darker red colors indicate higher likelihood. The numbers in the parentheses denote the number of in-training simulation steps used to train the model.
}
\label{fig:mesh_vis}
\vspace{-0.1in}
\includegraphics[width=0.95\linewidth]{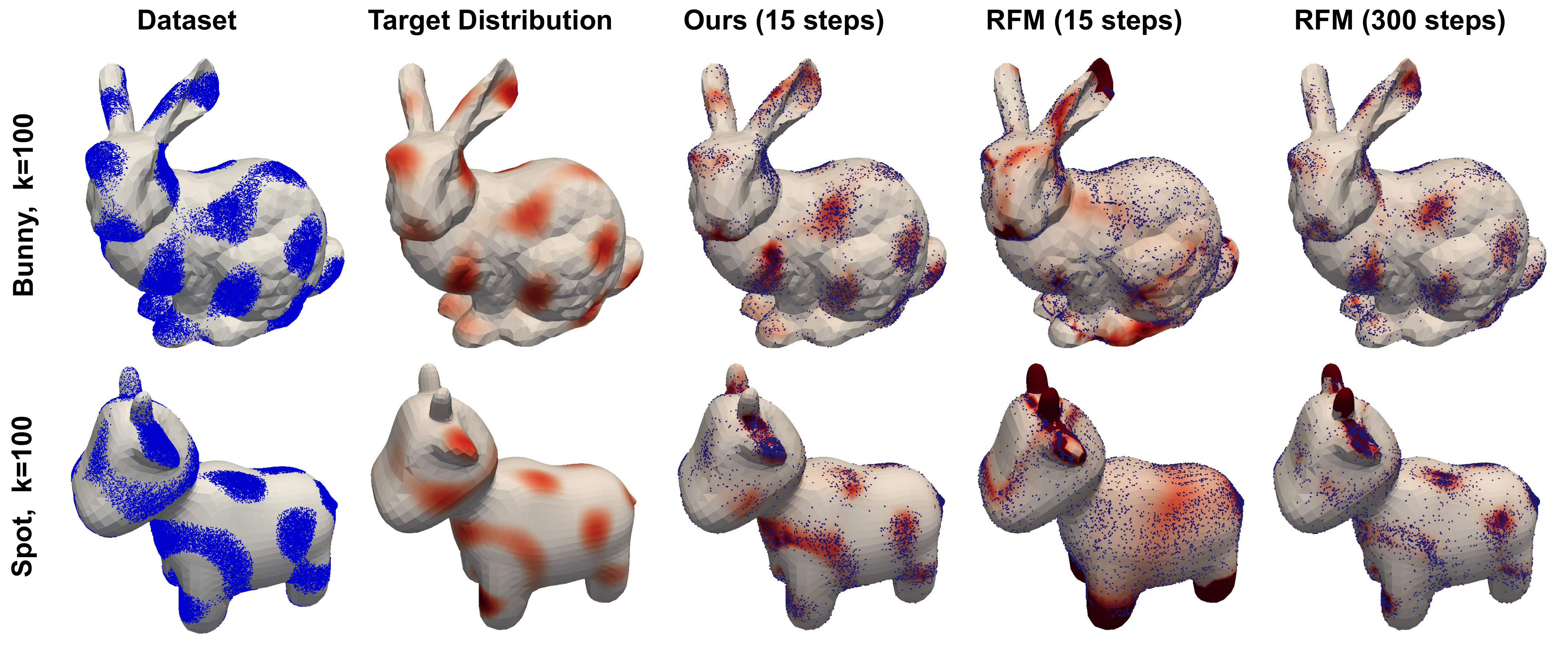}
\end{figure*}
%%%%%%%%%%%%%%%%%%%%%%%%%%%%%%%%%%%%%%%%
\begin{figure*}[!t]
\centering
\vspace{-0.15in}
\begin{minipage}{0.68\linewidth}
    \centering
\resizebox{\textwidth}{!}{
\renewcommand{\arraystretch}{1.0}
\renewcommand{\tabcolsep}{7pt}
\begin{tabular}{l @{\hskip1pt} c c c c c}
\toprule
     & \multirow{2}{*}{Steps} & \multicolumn{2}{c}{Stanford Bunny} & \multicolumn{2}{c}{Spot the Cow} \\
     & & $k=50$ & $k=100$ & $k=50$ & $k=100$ \\
\midrule
    RFM w/ Diff. Dist. & 300 & 1.48~\scriptsize{$\pm$ 0.01} & 1.53~\scriptsize{$\pm$ 0.01} & \textbf{0.95}~\scriptsize{$\pm$ 0.05} & 1.08~\scriptsize{$\pm$ 0.05} \\
    RFM w/ Bihar. Dist. & 300 & 1.55~\scriptsize{$\pm$ 0.01} & 1.49~\scriptsize{$\pm$ 0.01} & \textbf{1.08}~\scriptsize{$\pm$ 0.05} & 1.29~\scriptsize{$\pm$ 0.05} \\
\midrule
    Ours w/ Diff. Dist. & 15 & \textbf{1.42}~\scriptsize{$\pm$ 0.01} & \textbf{1.41}~\scriptsize{$\pm$ 0.00} &  \textbf{0.99}~\scriptsize{$\pm$ 0.03} & \textbf{0.97}~\scriptsize{$\pm$ 0.03} \\
    Ours w/ Bihar. Dist. & 15 & 1.55~\scriptsize{$\pm$ 0.02} & 1.45~\scriptsize{$\pm$ 0.01} & \textbf{1.09}~\scriptsize{$\pm$ 0.06} & \textbf{0.97}~\scriptsize{$\pm$ 0.02} \\
\bottomrule
\end{tabular}}
\end{minipage}
\hfill
\begin{minipage}{0.31\linewidth}
\resizebox{\textwidth}{!}{
\renewcommand{\arraystretch}{1.1}
\renewcommand{\tabcolsep}{6pt}
\begin{tabular}{l c c}
\toprule
     & Torus & Spot \\
\midrule
    RSGM (ISM) & 4.97 & - \\
    RSGM (SSM) & 1.20 & - \\
    RFM & 0.97 & 12.83 \\
\midrule
    Ours & 1.00 & \phantom{0}1.00 \\
\bottomrule
\end{tabular}}
\end{minipage}
\vspace{-0.05in}
\captionof{table}{\textbf{(Left) Test NLL results on mesh datasets}. We report the mean of 5 different runs. Best performance and its comparable results ($p>0.05$) from the t-test are highlighted. 
\textbf{(Right) Comparison of the training time}. We report the relative training time of the baselines with respect to ours on high-dimensional torus and Spot.
}
\label{fig:mesh_and_time}
\vspace{-0.05in}
\end{figure*}

%%%%%%%%%%%%%%%%%%%%%%%%%%%%%%%%%%%%%%%%

\subsection{Earth and Climate Science Datasets}
We first evaluate the generative models on real-world datasets living on the 2-dimensional sphere, which consists of earth and climate science events including volcanic eruptions~\citep{data_volcano}, earthquakes~\citep{data_earthquake}, floods~\citep{data_flood}, and wild fires~\citep{data_fire}. We use the LogBM, i.e. a mixture of Logarithm bridges, as the logarithm map is easy to compute on the sphere. 

Table~\ref{tab:earth} demonstrates that our method significantly outperforms the baselines on the Volcano and the Fire datasets which require high fidelity as the data are concentrated in specific regions. Ours constantly outperforms the Riemannian score-based model while matching the performance of RFM on the Earthquake and the Flood dataset. We visualize the generated samples and learned densities of our model in Figure~\ref{fig:earth_vis} showing that our method is capable of capturing the distribution on the sphere. 

We further compare the convergence of the generative processes measured by the geodesic distance in Figure~\ref{fig:earth_conv} where our generative process converges faster than that of the baselines. We observe that the prediction from our model (Eq.~\eqref{eq:prediction}) converges faster than that of RFM, verifying that ours is able to make more accurate predictions throughout the generation process.

\subsection{Protein Datasets}
We further experiment on protein datasets represented on $n$-dimensional torus from the torsion angles, consisting of 500 high-resolution proteins~\citep{lovell2003structure} and 113 selected RNA sequences~\cite{murray2003rna} preprocessed by \citet{huang2022rdm} (details in Appendix~\ref{app:exp:protein}). We additionally compare our method against the Mixture of Power Spherical (MoPS)~\citep{de2020power} which models the distribution as a mixture of power spherical distributions. We use the LogBM as the logarithm map is easy to compute on the torus. Table of Figure~\ref{fig:torus} demonstrates that ours outperforms or is on par with RDM while making marginal improvements over RFM, where the baselines are likely to be close to optimal. We provide the results of the other two protein datasets (General, Pre-Pro) in Table~\ref{tab:protein} showing comparable results with RFM. We visualize the learned density in Figure~\ref{fig:ramachandran} using Ramachandran plots where ours models the data distribution almost perfectly.

\subsection{High-Dimensional Tori \label{sec:exp:htori}}
We validate the scalability of our method using the synthetic data on high-dimensional tori. We follow \citet{debortoli2022rsgm} by creating a wrapped Gaussian distribution with a random mean and variance of 0.2 on $n$-dimensional tori where we compare the performance with RFM and RSGM trained via implicit score matching. To make a fair comparison, we use the same model architecture for all methods, where the total number of parameters for our models match that of the baselines. We describe the detailed setting in Appendix~\ref{app:exp:htori}. As shown in Figure~\ref{fig:torus}~(Right), ours constantly outperforms RSGM, especially in high dimensions. 
RSGM scales poorly with the dimensions due to the high variance in computing the stochastic divergence of the training objective. RFM also shows a significant drop in high dimensions which implies that the vector field could not be well-approximated with a limited number of parameters. On the other hand, ours is able to scale fairly well even for high dimensions as our training objective of Eq.~\eqref{eq:importance_sampling} does not require computation of divergence or any approximation. We observe that our method shows consistent performance without degradation for higher dimensions when using more parameters, scaling fairly well even to dimension $10^4$.

In particular, as shown in Table~\ref{fig:mesh_and_time} (Right), we achieve up to $\times$5 speedup in training compared to RSGM that uses implicit score matching (ISM), and also significantly faster than RSGM using ISM with the stochastic estimator (SSM). 
Our training time is comparable to Flow Matching which is simulation-free on tori.

\subsection{General Closed Manifolds \label{sec:exp:mesh}}
To validate our framework on general manifolds with non-trivial curvature, we evaluate modeling synthetic distributions on triangular meshes. Following \citet{chen2024rfm}, we construct the target distribution on Stanford Bunny~\citep{turk1994zippered} and Spot the Cow~\citep{crane2013robust} from the $k$-th eigenfunction of the mesh, which we provide detail in Appendix~\ref{app:exp:mesh}. 
We use SpecBM, i.e. a mixture of Spectral bridges, with using the diffusion distance~\citep{coifman2006diffusion} or the biharmonic distance~\citep{lipman2010biharmonic} for $d_w$. As visualized in Figure~\ref{fig:mesh_vis} and Figure~\ref{fig:mesh_vis_abl}, our method is able to fit the complex distributions using only 15 steps for the in-training simulation, while RFM completely fails when using a small number of in-training simulation steps. We show in Table~\ref{fig:mesh_and_time} that our method outperforms RFM while using only 5\% of in-training simulation steps, achieving $\times$12.8 speed up in training compared to RFM.

\subsection{Further Analysis \label{sec:exp:abl}}

\paragraph{Non-Compact Manifold}
We further validate that our framework can be applied to non-compact manifolds, in particular for spaces of negative curvature. We experiment on the synthetic distributions on a 2-dimensional hyperboloid modeled by a mixture of wrapped Gaussian distributions.
Figure~\ref{fig:hyperboloid} demonstrates that our method is capable of modeling the target distributions.

\paragraph{Time-scaled Training Objective}
We experimentally validate that the time-scaled objective of Eq.~\eqref{eq:importance_sampling} is crucial for learning the distribution, by comparing ours with a variant trained with uniform time distribution similar to Flow Matching. Table~\ref{tab:time_weight_abl} shows that the variant using uniform time distribution results in a significant drop in performance. This is because our time-scaled objective guarantees maximizing the likelihood of the generative model, whereas the variant using uniform time distribution does not.

\paragraph{Number of In-Training Simulation Steps}
We empirically demonstrate that our method can be trained using only 15 steps for the in-training simulation in Figures~\ref{fig:abl_num_steps} and \ref{fig:abl_num_steps_w}: We show in (a) and (c) that the trajectories of the mixture process simulated with 15 steps result in an almost similar distribution to the exact trajectories, which cannot be achieved with the one-way simulation as shown in (b). Especially, (d) demonstrates that using a small noise scale, resembling a deterministic process, requires a large number of simulation steps to obtain accurate trajectories, explaining the reason for the failure of RFM in Figure~\ref{fig:mesh_vis} and \ref{fig:mesh_vis_abl}.

\section{Conclusion}
In this work, we present Riemannian Diffusion Mixture, a new approach for learning generative diffusion processes on general manifolds.
We build the generation process using a mixture of bridge processes by designing the drift to be a weighted mean of tangent directions to the data distribution, which does not require approximating the heat kernel. We develop a highly scalable training scheme on general manifolds based on simple regression of the drifts which enables significantly faster training compared to previous diffusion models.
Our approach shows superior performance on diverse manifolds with a dramatically reduced number of in-training simulation steps.
We believe our work provides a promising direction for manifold diffusion models which could be applied to various scientific fields, for example, the design of proteins.

\paragraph{Impact Statement}
This paper presents work whose goal is to advance the field of deep generative models for data on non-Euclidean spaces. We believe that our work can enhance our understanding of diverse scientific fields including protein modeling and high-energy physics.

\paragraph{Acknowledgement}
This work was supported by Institute for Information \& communications Technology Promotion(IITP) grant funded by the Korea government(MSIT) (No.2019-0-00075 Artificial Intelligence Graduate School Program(KAIST)), Institute of Information \& communications Technology Planning \& Evaluation (IITP) grant funded by the Korea government(MSIT) (No.2022-0-00713) and the National Research Foundation of Korea(NRF) grant funded by the Korea government(MSIT) (No. RS-2023-00256259).

\bibliography{reference}
\bibliographystyle{icml2024}

\newpage
\appendix
\onecolumn

\begin{center}{\bf {\LARGE Appendix}}\end{center}

\section{Derivations \label{app:derivations}}

\subsection{Diffusion Process on Riemannian Manifold \label{app:derivations:bb}}
Brownian bridge process is a diffusion process described by the Brownian motion conditioned to fixed endpoints, which is induced by the following infinitesimal generator:
\begin{align}
    \frac{1}{2}\Delta_{\mathcal{M}} + \nabla\log p(\cdot, z, T-t) ,
\end{align}
where $p$ denotes the transition density of the Brownian motion, i.e. the heat kernel defined on $\mathcal{M}$, and $z$ denotes the fixed endpoint. Thus Brownian bridge process can be modeled by the following SDE:
\begin{align}
    \mathbb{Q}^{z}_{bb} : \mathrm{d}\bm{X}_t = \nabla_{\!\bm{X}_t}\! \log p_{\mathcal{M}}(\bm{X}_t, z, T\!-\!t)\mathrm{d}t + \mathrm{d}\mathbf{B}^{\mathcal{M}}_t .
\end{align}
We refer the readers to \citet{hsu2002stochastic} for a formal definition of the Brownian bridge process. The theoretical properties of Brownian bridge have been studied in previous works~\citep{hsu1990brownian, driver1994cameron}.

One can construct a diffusion process on $\mathcal{M}$ that converges to a stationary distribution described by the Langevin dynamics as follows: 
\begin{align}
    \mathrm{d}\bm{X}_t = -\frac{1}{2}\nabla_{\bm{X}_t}U(\bm{X}_t) \mathrm{d}t + \mathrm{d}\mathbf{B}^{\mathcal{M}}_t,
\end{align}
where the terminal distribution satisfies $\mathrm{d}p(x)/\mathrm{d}\text{vol}_x\propto e^{-U(x)}$~\citep{durmus2016high} for a potential function $U$. For example, $U(x) = d_g(x, \mu)^2 / (2\gamma^2) + \log \|D \exp^{-1}_{\mu}(x)\|$ results in a stationary distribution equal to the wrapped Gaussian distribution with an arbitrary mean location $mu\in\mathcal{M}$, where $d_g$ denotes the geodesic distance.

\subsection{Logarithm Bridge Process \label{app:derivations:lbp}}
Here we show that the Logarithm bridge of Eq.~\eqref{eq:logarithm_bridge} describes a diffusion process that converges to an endpoint. For the notational simplicity, we omit the subscript $\mathcal{M}$ for the heat kernel on $\mathcal{M}$.
First, we derive that the following simplified process is a bridge process with an endpoint $z$:
\begin{align}
    \mathbb{Q}^{z}_{log} : \mathrm{d}\bm{X}_t = \frac{1}{T-t} \exp^{\scalebox{0.75}[1.0]{-}1}_{\!\bm{X}_t}\!(x) \mathrm{d}t + \mathrm{d}\mathbf{B}^{\mathcal{M}}_t .
\label{eq:simple_logarithm_bridge}
\end{align}

For any pair of points $x$ and $y$ on $\mathcal{M}$, the following short-time asymptotic of the heat kernel holds~(Theorem 5.2.1 of \citep{hsu2002stochastic}):
\begin{align}
    \lim_{t\rightarrow0}t\log p(x,y,t) = -\frac{d_g(x,y)^2}{2},
\end{align}
where $d_g(\cdot,\cdot)$ is a geodesic distance defined on $\mathcal{M}$. Further leveraging the identity from Proposition 6 of \citet{mccann2001polar}, we obtain the following result:
\begin{align}
    \lim_{t\rightarrow0}t\nabla_x\log p(x,y,t) = -\frac{1}{2}\nabla_x d_g(x,y)^2 = \exp^{\scalebox{0.75}[1.0]{-}1}_x(y) .
\label{eq:short_time_asymptotic}
\end{align}
Furthermore, we have an upper bound for the gradient of the logarithmic heat kernel for any pair of points $x,y\in\mathcal{M}$ and $t\in [0,T]$ as follws~(Theorem 5.5.3 of \citep{hsu2002stochastic}):
\begin{align}
    \Big\| \nabla_x\log p(x, y, t) \Big\|_{\mathcal{M}} \leq C \left[ \frac{d_g(x, y)}{t} + \frac{1}{\sqrt{t}} \right] ,
\label{eq:log_heat_kernel_upper_bound}
\end{align}
where $C$ is a constant. From Eq.~\eqref{eq:log_heat_kernel_upper_bound}, we have the following bound:
\begin{align}
    &\left\| \frac{\exp^{\scalebox{0.75}[1.0]{-}1}_x(z)}{T-t} - \nabla_x\log p\left(x, z, T\!-\!t \right) \right\|_{\mathcal{M}} \\
    &= \mathbbm{1}_{\{t>T-\epsilon\}} \cdot \left\| \frac{\exp^{\scalebox{0.75}[1.0]{-}1}_x(z)}{T-t} - \nabla_x\log p\left(x, z, T\!-\!t \right) \right\|_{\mathcal{M}} \label{eq:log_derivation_delta_term} 
    + \mathbbm{1}_{\{t\leq T-\epsilon\}} \cdot \left\| \frac{\exp^{\scalebox{0.75}[1.0]{-}1}_x(z)}{T-t} - \nabla_x\log p\left(x, z, T\!-\!t \right) \right\|_{\mathcal{M}} \\[5pt]
    &\leq \delta(\epsilon) + \frac{1}{\epsilon} \left\| \exp^{\scalebox{0.75}[1.0]{-}1}_x(z) \right\|_{\mathcal{M}} + \left\| \nabla_x\log p(x, z, T\!-\!t) \right\|_{\mathcal{M}} \\[5pt]
    &\leq \delta(\epsilon) + \frac{1}{\epsilon} d_g(x, z) + C \left[ \frac{d_g(x, z)}{\epsilon} + \frac{1}{\sqrt{\epsilon}} \right] ,
\end{align}
which holds for every $\epsilon>0$ where $\delta(t)$ denotes the first term of Eq.~\eqref{eq:log_derivation_delta_term}. Since $\delta(\epsilon)\rightarrow 0$ as $\epsilon\rightarrow 0$ from the results of Eq.~\eqref{eq:short_time_asymptotic}, we obtain the following bound:
\begin{align}
    \left\| \frac{\exp^{\scalebox{0.75}[1.0]{-}1}_x(z)}{T-t} - \nabla_x\log p\left(x, z, T\!-\!t \right) \right\|_{\mathcal{M}} < \infty .
\end{align}
Finally, using the Girsanov theorem for $\mathbb{Q}^{z}_{bb}$ and $\mathbb{Q}^{z}_{log}$, we obtain the following result:
\begin{align}
    D_{KL}\left(\mathbb{Q}^{z}_{bb} \| \mathbb{Q}^{z}_{log} \right) 
    = \frac{1}{2}\mathbb{E}_{\bm{X}\sim\mathbb{Q}^{z}_{bb}}\left[\int^T_0 \left\| \frac{\exp^{\scalebox{0.75}[1.0]{-}1}_{\bm{X}_t}(z)}{T-t} - \nabla_x\log p\left(\bm{X}_t, z, T\!-\!t \right) \right\|^2_{\mathcal{M}} \mathrm{d}t \right] < \infty,
\end{align}
which implies that the Brownian bridge and the Logarithm bridge have the same support. Since the Brownian bridge converges to $z$ by definition, we can conclude that the Logarithm bridge also converges to a fixed endpoint $z$.

The Logarithm bridge process presented in Eq.~\eqref{eq:logarithm_bridge} is the result of using the change of time~\citep{oksendal2003sde} on Eq.~\eqref{eq:simple_logarithm_bridge} with respect to the rescaled time $\tau(t)\coloneqq\int^t_0 \sigma^2_s\mathrm{d}s$. 
Additionally, similar to the Euclidean bridge processes introduced in \citet{wu2022bridge}, the derivation of our Logarithm bridge process provides a more general form of bridge processes on the manifold:
\begin{align}
     \mathrm{d}\bm{X}_t = \left[ \frac{\sigma_t^2}{\tau_T-\tau_t} \exp^{\scalebox{0.75}[1.0]{-}1}_{\!\bm{X}_t}\!(z) + \sigma_t \nabla_{\bm{X}_t}U(\bm{X}_t,t) \right] \mathrm{d}t + \sigma_t\mathrm{d}\mathbf{B}^{\mathcal{M}}_t, 
\label{eq:general_logarithm_bridge}
\end{align}
for a scalar function $U$ that satisfies $\mathbb{E}_{\bm{X}\sim\mathbb{Q}^{z, bb}}\int^T_0 \| \nabla U(\bm{X}_t,t) \|^2_{\mathcal{M}}<\infty$ which is sufficient for bounded functions $U$. 

\subsection{Other Bridge Processes} \label{app:derivations:general_bridge}

\paragraph{Semi-classical Brownian bridge}
A semi-classical Brownian bridge introduced by \citep{elworthy1981classical}, also known as Brownian Riemannian bridge, is a bridge process defined by the infinitesimal generator
\begin{align}
    \frac{1}{2}\Delta_{\mathcal{M}} + \nabla\log k(\cdot, z, T-t) ,
\end{align}
where $k$ is given by the Jacobian determinant of the exponential map at y as follows:
\begin{align}
    k(x, y, t) = \frac{1}{(2\pi t)^{n/2}} e^{-\frac{d_{g}^2(x,y)}{2t}} \lvert \text{det}D_{\exp^{\scalebox{0.75}[1.0]{-}1}_{y}(x)}\exp_y \rvert^{-1/2} .
\end{align}

\paragraph{Fermi bridge}
Previous works~\citep{thompson2015submanifold, thompson2018brownian} introduce a general family of bridge processes, namely the Fermi bridge, that describes diffusion processes conditioned to a submanifold $N$, which can be defined by the infinitesimal generator as follows:
\begin{align}
    \frac{1}{2}\Delta_{\mathcal{M}} - \frac{r_N}{T-t}\frac{\partial}{\partial r_N} ,
\end{align}
where $r_N(\cdot)\!\coloneqq d(\cdot,N)$ is the distance function to the submanifold $N$ and $\partial / \partial r_N$ denotes differentiation in radial direction. 
Although the Logarithm bridge can be derived from the Fermi bridge by constraining $N$ to a single point, to the best of our knowledge, the Logarithm bridge was not studied in previous literature, and further leveraging the Logarithm bridge process in the context of generative modeling is our novel contribution.

\subsection{Diffusion Mixture on Riemannian Manifold \label{app:derivations:mix}}

\paragraph{Diffusion Mixture Representation}
We extend the diffusion mixture representation~\citep{peluchetti2021mixture} to the Riemannian setting. We start with the statement of the diffusion mixture representation: 
Let $\{\mathbb{Q}^{\lambda}:\lambda\in\Lambda\}$ be a collection of diffusion processes on $\mathcal{M}$ modeled by the following SDEs:
\begin{align}
    \mathbb{Q}^{\lambda} : \mathrm{d}\bm{X}^{\lambda}_t =  \eta^{\lambda}(\bm{X}^{\lambda}_t, t) \mathrm{d}t + \sigma^{\lambda}_t \mathrm{d}\mathbf{B}^{\lambda}_t \;, \;\; \bm{X}^{\lambda}_0\sim p^{\lambda}_0,
\end{align}
where $\mathbf{B}^{\lambda}_t$ are independent Brownian motions on $\mathcal{M}$ and $p^{\lambda}_0$ denotes the initial distributions of $\mathbb{Q}^{\lambda}$. 
Denoting the marginal distribution of $\mathbb{Q}^{\lambda}$ as $p^{\lambda}_t$ and a mixing distribution $\mathcal{L}$ on $\Lambda$, consider the following density and the distribution defined as the mixture of $p^{\lambda}_t$ and $p^{\lambda}_0$, respectively, as follows:
\begin{align}
    p_t(x) = \int p^{\lambda}_t(x) \mathcal{L}(\mathrm{d}\text{vol}_x) \;,\;\;  p_0(x) = \int p^{\lambda}_0(x) \mathcal{L}(\mathrm{d}\text{vol}_x) .
\end{align}
Then there exists a diffusion process on $\mathcal{M}$ such that its marginal density is equal to $p_t$ with the initial distribution given as $p_0$, described by the following SDE:
\begin{align}
    \mathbb{Q}^{\ast} : \mathrm{d}\bm{X}_t =  \eta(\bm{X}_t, t) \mathrm{d}t + \sigma_t \mathrm{d}\mathbf{B}^{\mathcal{M}}_t \;,\;\; \bm{X}_0\sim p_0 ,
    \label{eq:diffusion_mixture_process}
\end{align}
where the drift $\eta_t$ and the diffusion coefficient $\sigma_t$ satisfy the following:
\begin{align}
    \eta(x, t) = \int \eta^{\lambda}(x, t) \frac{p^{\lambda}_t(x)}{p_t(x)} \mathcal{L}(\mathrm{d}\text{vol}_x) \;, \;\; \sigma^2_t = \int (\sigma^{\lambda}_t)^2 \frac{p^{\lambda}_t(x)}{p_t(x)} \mathcal{L}(\mathrm{d}\text{vol}_x) .
    \label{eq:mixture_condition}
\end{align}

The proof for the Riemannian setting extends that of the Euclidean case, where we leverage the Fokker-Planck equation to characterize the marginal density. 
From the condition of Eq.~\eqref{eq:mixture_condition}, we can derive the following:
\begin{align}
    &\frac{\partial p_t(x)}{\partial t} = \int \frac{\partial}{\partial t} p^{\lambda}_t(x) \mathcal{L}(\mathrm{d}\text{vol}_x) \\[5pt]
    &= \int \left[ -\text{div}\Big(p^{\lambda}_t(x) \eta^{\lambda}(x, t) \Big) + \frac{1}{2} (\sigma^{\lambda}_t)^2\Delta_{\mathcal{M}} p^{\lambda}_t(x)  \right] \mathcal{L}(\mathrm{d}\text{vol}_x) \\[5pt]
    &= -\text{div} \left( p_t(x)\int \eta^{\lambda}(x,t)\frac{p^{\lambda}_t(x)}{p_t(x)} \mathcal{L}(\mathrm{d}\text{vol}_x) \right) 
    +\frac{1}{2}\Delta_{\mathcal{M}} \left( p_t(x) \int (\sigma^{\lambda}_t)^2 \frac{p^{\lambda}_t(x)}{p_t(x)}\mathcal{L}(\mathrm{d}\text{vol}_x) \right)  \\[5pt]
    &= -\text{div} \Big(  p_t(x)\eta(x,t) \Big) + \frac{1}{2}\sigma^2_t\Delta_{\mathcal{M}} p_t(x) ,
    \label{eq:fokker_planck_mixture}
\end{align}
where the second equality is derived from the Fokker-Planck equation with respect to the process $\mathbb{Q}^{\lambda}$.
Since Eq.~\eqref{eq:fokker_planck_mixture} corresponds to the Fokker-Planck equation with respect to the mixture process, we can conclude that $p_t$ is the marginal density of the mixture process.

\paragraph{Generative Process}
Now, we are ready to derive our Riemannian Diffusion Mixture in Eq.~\eqref{eq:mixture_process}. 
Using the diffusion mixture representation, we can derive the mixture of a collection of the bridge processes $\{ \mathbb{Q}^{x,z}:x\!\sim\!\Gamma, z\!\sim\!\Pi \}$ as follows: 
\begin{align}
    \mathbb{Q}^{\Pi}: \mathrm{d}\bm{X}_t 
    &= \left[ \int \int \eta^{z}(\bm{X}_t,t) \frac{p^{x,z}(\bm{X}_t)}{p_t(\bm{X}_t)} \Gamma(\mathrm{d}\text{vol}_x) \Pi(\mathrm{d}\text{vol}_z) \right] \mathrm{d}t + \sigma_t\mathrm{d}\mathbf{B}^{\mathcal{M}}_t \\
    &= \left[ \int \eta^{z}(\bm{X}_t,t) \frac{p^{z}(\bm{X}_t)}{p_t(\bm{X}_t)} \Pi(\mathrm{d}\text{vol}_z) \right] \mathrm{d}t + \sigma_t\mathrm{d}\mathbf{B}^{\mathcal{M}}_t
    \;,\; \bm{X}_0\!\sim\!\Gamma,
\end{align}
where $\eta^{z}$ denotes the drift of the bridge processes $\mathbb{Q}^{x,z}$ for $x\!\sim\!\Gamma$ and $p^z_t(\cdot) = \int p^{x,z}_t(\cdot) \Gamma(\mathrm{d}\text{vol}_x)$.

\subsection{Probability Flow ODE \label{app:derivations:pode}}
Here we provide the derivation of the probability flow of the mixture process $\mathbb{Q}_f$ by considering its time-reversed process $\mathbb{Q}_b$ in two different perspectives. First, the time-reversed process $\mathbb{Q}_b$ corresponds to a mixture process built from the collection of time-reversed bridge processes from $\Pi$ to $\Gamma$, which can be modeled by the following SDE:
\begin{align}
    \mathrm{d}\overbar{\bm{X}}_t = \frac{\sigma_{T\!-\!t}^2}{\tau_T-\tau_{T\!-\!t}} \left[ \int \eta^{z}_b(\overbar{\bm{X}}_t, t) \frac{p^{z}_t(\overbar{\bm{X}}_t)}{p_t(\overbar{\bm{X}}_t)}\, \Gamma(\mathrm{d}\text{vol}_z) \right] \mathrm{d}t  + \sigma_{T\!-\!t}\mathrm{d}\mathbf{B}^{\mathcal{M}}_t \;,\;\; \overbar{\bm{X}}_0\sim\Pi,
\label{eq:reversed_mixture_process}
\end{align}
where $\tau(t) \coloneqq \int^t_0 \! \sigma_s^2\mathrm{d}s$ as defined in Eq.~\eqref{eq:logarithm_bridge} and $\eta^{z}_b$ denotes the drift of the bridge process $\mathbb{Q}_b(\cdot|\overbar{\bm{X}}_T=z)$. 
On the other hand, $\mathbb{Q}_b$ can be derived in terms of the score function from Theorem 3.1 of \citet{debortoli2022rsgm} as follows:
\begin{align}
    \mathrm{d}\overbar{\bm{X}}_t = \Big[ -\eta_f(\overbar{\bm{X}}_t, t) + \sigma_{T\!-\!t}^2 \nabla_{\overbar{\bm{X}}_t}\log p_t(\overbar{\bm{X}}_t) \Big] \mathrm{d}t + \sigma_{T\!-\!t}\mathrm{d}\mathbf{B}^{\mathcal{M}}_t .
\label{eq:reversed_mixture_process_score}
\end{align}
Therefore, we can obtain the score function in terms of the drifts of $\mathbb{Q}_f$ and $\mathbb{Q}_b$ as follows:
\begin{align}
    \nabla\log p_t(\bm{X}_t) = \left( \eta_f(\bm{X}_t,t) + \eta_b(\bm{X}_t,T\!-\!t) \right) / \sigma_t^2 ,
\end{align}
and as a result, the probability flow associated with the mixture process $\mathbb{Q}_f$ is obtained by the following ODE:
\begin{align}
    \frac{\mathrm{d}}{\mathrm{d}t} \bm{Y}_t 
    &= \Big( \eta_f(\bm{Y}_t,t) - \frac{1}{2}\nabla\log p_t(\bm{Y}_t) \Big) = \frac{1}{2} \Big( \eta_f(\bm{Y}_t,t) - \eta_b(\bm{Y}_t,T\!-\!t) \Big) \;,\;\; \bm{Y}_0\sim \Gamma .
    \label{eq:probability_flow_appendix}
\end{align}
When $\mathcal{M}$ is a Euclidean space, our derived probability flow of Eq.~\eqref{eq:probability_flow_appendix} corresponds to the probability flow for Schr\"odinger bridges~\citep{debortoli2021bridge,chen2022fbsde,shi2023dsbm}.

Finally, using Proposition 2 of \citet{mathieu2020rcnf}, we can compute the likelihood of the probability flow as follows:
\begin{align}
    \log p_T(\bm{Y}_T) = \log p_0(\bm{Y}_0) + \frac{1}{2} \int^{T}_0 \text{div}\Big( \eta_f(\bm{Y}_t,t) - \eta_b(\bm{Y}_{t},T\!-\!t) \Big) \mathrm{d}t .
\end{align}

It is worth noting that the probability flow of the mixture process is different from the continuous flows used in previous works. 
This is due to the difference in the marginal densities that are characterized by different laws:
By construction, the marginal density of the probability flow is equal to the marginal density of the associated mixture process which is described by the Fokker-Planck equation:
\begin{align}
    \frac{\partial p_t(z)}{\partial t} = -\text{div}\Big( \eta_f(z, t)p_t(z) \Big) + \frac{1}{2}\sigma_t^2\Delta_{\mathcal{M}}p_t(z),
\end{align}
whereas the marginal density of a deterministic process is described by the transportation equation:
\begin{align}
     \frac{\partial \tilde{p}_t(z)}{\partial t} = -\text{div} \Big( \eta_{\text{CNF}}(z, t)\tilde{p}_t \Big) .
\end{align}

\subsection{Bridge Matching on Riemannian Manifold \label{app:derivations:bridge_matching}}
We first derive the KL divergence between a diffusion process $\mathbb{Q}: \mathrm{d}\bm{Z}_t \!=\! \eta(\bm{Z}_t,t)\mathrm{d}t + \nu_t \mathrm{d}B^{\mathcal{M}}_t$ with terminal distribution $\mathbb{Q}_T$ and its parameterized process $\mathbb{P}^{\psi}: \mathrm{d}\bm{Z}_t = s^{\psi}\!(\bm{Z}_t,t)\mathrm{d}t + \nu_t\mathrm{d}B^{\mathcal{M}}_t$ by leveraging the Girsanov theorem as follows:
\begin{align}
    D_{KL}(\mathbb{Q}_T \| \mathbb{P}^{\psi}_T) &\leq D_{KL}(\mathbb{Q} \| \mathbb{P}^{\psi})
    = \mathbb{E}_{\substack{z\sim \mathbb{Q}_T, \\ \bm{Z}\sim\mathbb{Q}^z}} \!\! \left[ \log \frac{\mathrm{d}\mathbb{Q}^{z}}{\mathrm{d}\mathbb{P}^{\psi}}(\bm{Z}) + \log \frac{\mathrm{d}\mathbb{Q}}{\mathrm{d}\mathbb{Q}^{z}}(\bm{Z}) \right] \\[5pt]
    &= \mathbb{E}_{z\sim \mathbb{Q}_T}\Big[ D_{KL}(\mathbb{Q}^z  \|  \mathbb{P}^{\psi} ) \Big] + C_1 \\[5pt]
    &= \mathbb{E}_{\substack{z\sim \mathbb{Q}_T, \\ \bm{Z}\sim\mathbb{Q}^z}} \!\! \left[ \frac{1}{2}\! \int^T_0 \!\! \left\| \nu_t^{-1} \Big(\bm{s}^{\psi}(\bm{Z}_t,t) - \eta^{z}(\bm{Z}_t,t) \Big) \right\|^2_{\mathcal{M}} \!\! \mathrm{d}t  \right] + C_2 ,
\label{eq:diffusion_kl}
\end{align}
where $\mathbb{Q}_T$ and $\mathbb{P}^{\psi}_T$ denotes the terminal distributions of $\mathbb{Q}$ and $\mathbb{P}^{\psi}$ respectively, $\mathbb{Q}^z$ and $\eta^{z}$ denotes the process $\mathbb{Q}(\cdot|\bm{Z}_T\!=\!z)$ and its drift, and $C$ is a constant.
The first inequality is from the data processing inequality. 

Using the result of Eq.~\eqref{eq:diffusion_kl} and leveraging the fact that the time-reversed process of the mixture process $\mathbb{Q}_f$ is also a mixture process of time-reversed bridge processes (Eq.~\eqref{eq:reversed_mixture_process}), the models $\bm{s}^{\theta}_f$ and $\bm{s}^{\phi}_b$ can be trained to approximate the drifts $\eta_f$ and $\eta_b$, respectively, with the following objectives:
\begin{align}
    \mathcal{L}_f(\theta) = \mathbb{E}_{\substack{x\sim \Pi, \\ \bm{X}\sim\mathbb{Q}^{x}_f}} &\left[ \frac{1}{2}\int^T_0 \left\| \sigma_t^{-1} \Big(\bm{s}^{\theta}_f(\bm{X}_t,t) - \eta^{x}_f(\bm{X}_t,t) \Big) \right\|^2_{\mathcal{M}} \mathrm{d}t\right] , \\
    \mathcal{L}_b(\phi) = \mathbb{E}_{\substack{y\sim \Gamma, \\ \overbar{\bm{X}}\sim\mathbb{Q}^{y}_b}} &\left[ \frac{1}{2}\int^T_0 \left\| \sigma_{T\!-\!t}^{-1} \Big(\bm{s}^{\phi}_b(\overbar{\bm{X}}_t,t) - \eta^{y}_b(\overbar{\bm{X}}_t,t) \Big) \right\|^2_{\mathcal{M}} \mathrm{d}t\right] .
\label{eq:separate_training}
\end{align}
Instead of separately training the models by simulating two different bridge processes $\mathbb{Q}^{x}_f$ and $\mathbb{Q}^{y}_b$, we can train the models simultaneously by simulating a single bridge process $\mathbb{Q}^{x,y}$ with a fixed starting point $x$ and endpoint $y$, i.e., $\mathbb{Q}_f(\cdot|\bm{X}_0=x,\bm{X}_T=y)$, reducing the computational cost for the simulation in half. Furthermore, when simulating $\mathbb{Q}^{x,y}$, we introduce a two-way approach to obtaining $\bm{Z}_t\sim\mathbb{Q}^{x,y}$: simulating $\mathbb{Q}^{x,y}$ from time $0$ to $t$ if $t<T/2$, and otherwise simulating from time $T$ to $t$.
Altogether, we obtain the following loss as presented in Eq.~\eqref{eq:twoway_bridge_matching}:
\begin{align}
    \mathbb{E}_{\substack{(x,y)\sim (\Pi,\Gamma), \\ \bm{Z}\sim\mathbb{Q}^{x,y}}} \frac{1}{2} \!\int^T_0 \!\!\sigma^{-2}_t
    \Bigg[ 
    \Big\| \bm{s}^{\theta}_f(\bm{Z}_t,t) - \eta^x_f(\bm{Z}_t,t)  \Big\|^2_{\mathcal{M}} \!\!+ \Big\| \bm{s}^{\phi}_b(\bm{Z}_t,T\!\!-\!t) - \eta^y_b(\bm{Z}_t,T\!\!-\!t) \Big\|^2_{\mathcal{M}} 
    \Bigg] \mathrm{d}t.
\end{align}
Furthermore, by leveraging an importance sampling with a proposal distribution $q(t)\propto \sigma_t^{-2}$, we obtain the time-scaled two-way bridge matching as follows:
\begin{align}
     \mathbb{E}_{\substack{(x,y)\sim (\Pi,\Gamma), \\ t \sim q}}
    \mathbb{E}_{\bm{Z}_t \sim \mathbb{Q}^{x,y}} 
    \Bigg[ 
    \Big\|  \bm{s}^{\theta}_f(\bm{Z}_t,t) - \eta^x_f(\bm{Z}_t,t) \Big\|^2_{\mathcal{M}} \!\! + \Big\| \bm{s}^{\phi}_b(\bm{Z}_t,T\!-\!t) - \eta^y_b(\bm{Z}_t,T\!-\!t)  \Big\|^2_{\mathcal{M}} \Bigg] .
\end{align} 
We note that while the idea of the importance sampling for the time distribution was also used in \citet{huang2022rdm}, our approach leverages a simple and easy-to-sample proposal distribution $q$, which is effective in stabilizing the training and improving the generation quality, without the need for additional computation or training time.

\subsection{Other Relevant Works \label{app:derivations:comparison}}
Here we further discuss relevant works on manifold diffusion models, extending Section~\ref{sec:related_work}. 
\citet{thornton2022rdsb} extends Diffusion Schr\"odinger Bridge to the manifold setting, which aims to find the forward and backward processes between distributions that minimize the KL divergence to the Brownian motion. However, \citet{thornton2022rdsb} uses Iterative Proportional Fitting to alternatively train the models that require computing the divergence for numerous iterations, which is computationally expensive compared to our divergence-free two-way bridge matching which can train the models simultaneously. 
Recently, \citet{lou2023scaling} introduced practical improvements for Riemannian diffusion models based on a refined estimator for the heat kernel on Riemannian symmetric spaces. For our framework, improved heat kernel estimators can be used to construct a mixture of Brownian bridges, instead of Logarithm bridges or Spectral bridges, but we leave it as future work.
Furthermore, \citet{bose2023se} introduces a stochastic version of Flow Matching~\citep{lipman2023flowmatching} in the SO(3) group, which can be considered a special case of our Logarithm bridge applied to SO(3).
Other line of works focus on specific geometries such as SO(3)~\citep{leach2022so3}, SE(3)~\citep{yim2023se3,urain2023se}, and product of tori~\citep{jing2022torsion}, or constrained manifolds~\citep{fishman2023constrained} defined by set of inequality constraints.

%%%%%%%%%%%%%%%%%%%%%%%%%%%%%%%%%%%%%%%%%
\begin{figure}[!t]
    \centering
    \includegraphics[width=\linewidth]{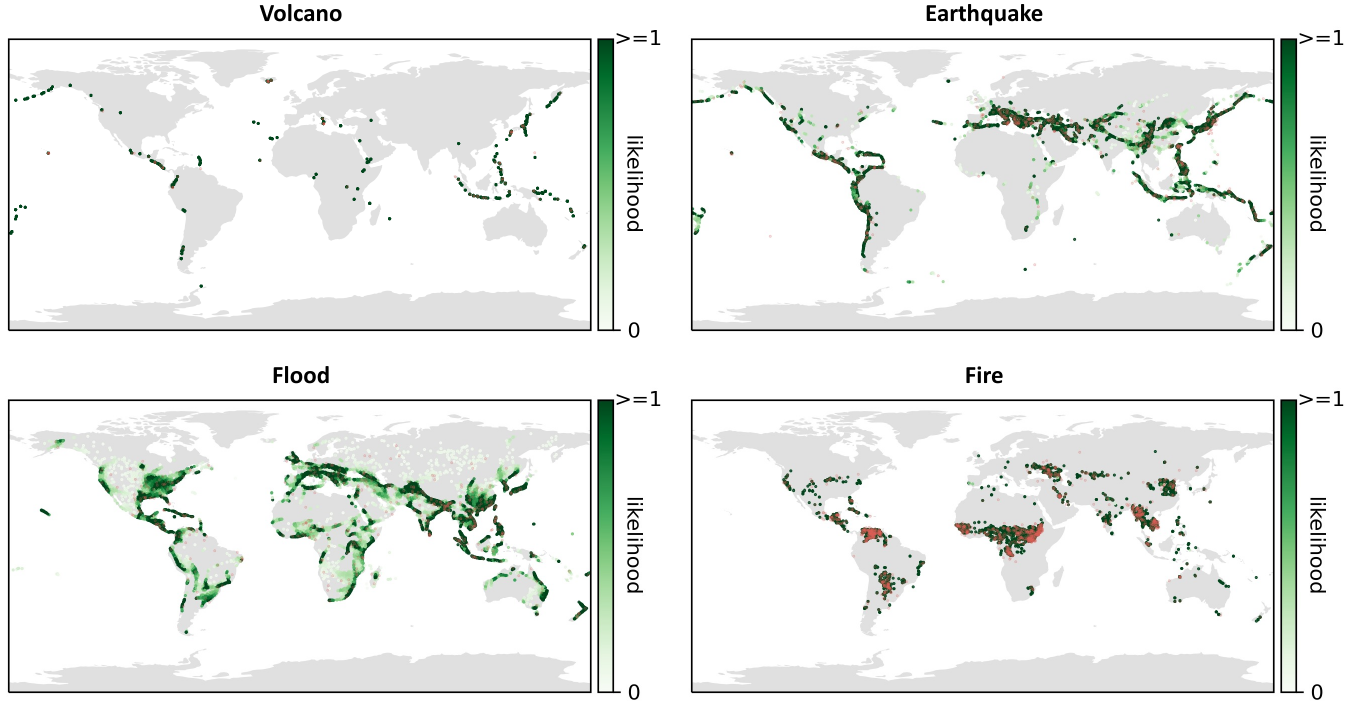}
\vspace{-0.15in}
    \caption{\textbf{Visualization of the generated samples and learned density} of our model on earth and climate science datasets. Red dots denote samples from the test set and green dots denote the generated samples. Darker green colors denote a higher likelihood modeled by our approach.}
    \label{fig:earth_vis}
\end{figure}
%%%%%%%%%%%%%%%%%%%%%%%%%%%%%%%%%%%%%%%%%
\begin{figure}[!t]
\vspace{-0.1in}
    \centering
\begin{minipage}{0.48\linewidth}
    \centering
    \includegraphics[width=1.0\linewidth]{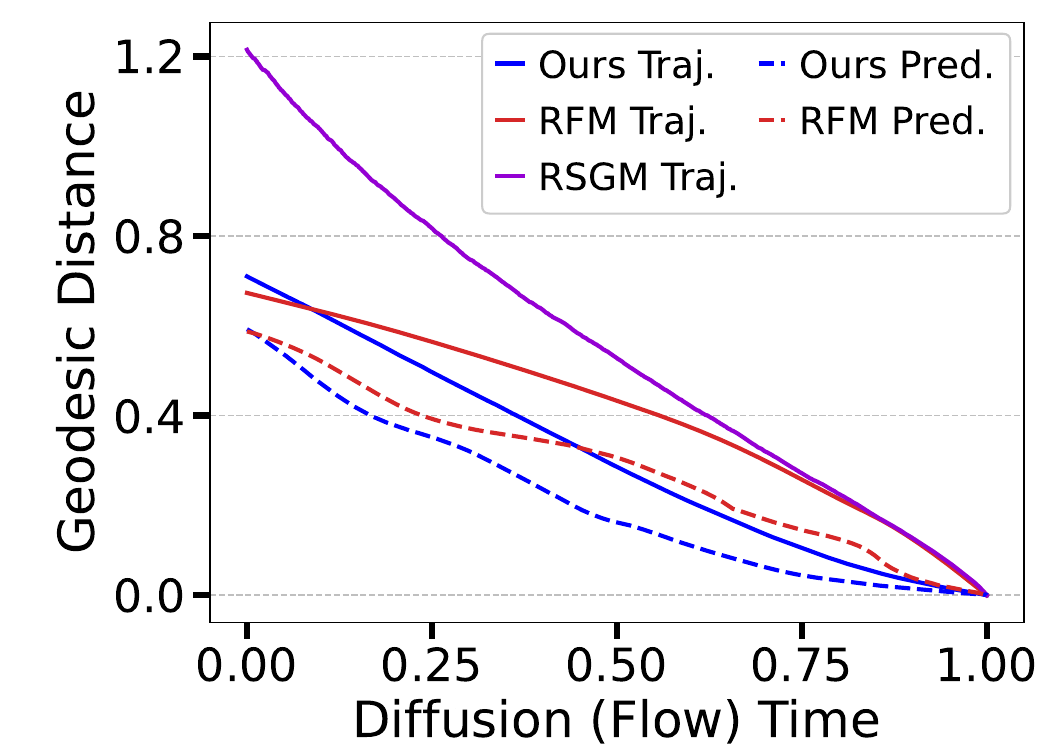}
\vspace{-0.25in}
    \caption{\textbf{Convergence of the generative processes} on the Volcano dataset. We compare the geodesic distance between the samples of the trajectory (Traj.) and the final results. We further compare the convergence of predictions (Pred.) made by ours and that of RFM. }
    \label{fig:earth_conv}
\end{minipage}
\hfill
\begin{minipage}{0.5\linewidth}
% \vspace{-0.075in}
\centering
    \resizebox{1.0\textwidth}{!}{
\renewcommand{\arraystretch}{1.0}
\renewcommand{\tabcolsep}{14pt}
\begin{tabular}{l c c}
\toprule
     Ours & Volcano & Flood \\
\midrule
    Uniform & -5.37~\scriptsize{$\pm$ 0.67} & 0.67~\scriptsize{$\pm$ 0.14} \\
    Time-scaled & \textbf{-9.52}~\scriptsize{$\pm$ 0.87} & \textbf{0.42}~\scriptsize{$\pm$ 0.08} \\
\bottomrule
\end{tabular}}
\vspace{-0.1in}
\caption{\textbf{Ablation study on the time-scaled training objective}. We report the test NLL of our method (Time-scaled) against a variant trained with uniformly distributed time (Uniform), instead of the time-scaled distribution $q$ in Eq.~\eqref{eq:importance_sampling}.
}
\label{tab:time_weight_abl}
\vspace{0.15in}
    \resizebox{1.0\textwidth}{!}{
\renewcommand{\arraystretch}{1.0}
\renewcommand{\tabcolsep}{10pt}
\begin{tabular}{l c c}
\toprule
     Ours & Earthquake & Proline \\
\midrule
    Different samples & -0.29~\scriptsize{$\pm$ 0.08} & 0.14~\scriptsize{$\pm$ 0.025} \\
    Same samples & -0.30~\scriptsize{$\pm$ 0.06} & 0.14~\scriptsize{$\pm$ 0.027} \\
\bottomrule
\end{tabular}}
\vspace{-0.1in}
\caption{\textbf{Ablation study on the two-way bridge matching}. We show that sampling $\bm{Z}_t$ from a single bridge process (denoted as Same samples) as proposed in our work does not introduce additional variance by comparing the performance with the variant of our method which samples different $\bm{Z}_t$ for each forward and backward direction of the bridge process (denoted as Different samples).
}
\label{tab:forward_backward_variance}
\end{minipage}
\vspace{-0.1in}
\end{figure}
%%%%%%%%%%%%%%%%%%%%%%%%%%%%%%%%%%%%%%%%%

%%%%%%%%%%%%%%%%%%%%%%%%%%%%%%%%%%%%%%%%%
\begin{table*}[t]
\caption{\textbf{Test NLL results on protein datasets}. We report the mean of 5 different runs with different data splits. Best performance and its comparable results ($p>0.05$) from the t-test are highlighted in bold. 
% The baseline results are taken from \citet{chen2024rfm}.
}
\label{tab:protein}
\vspace{-0.05in}
\centering
    \resizebox{\textwidth}{!}{
    \renewcommand{\arraystretch}{1.1}
    \renewcommand{\tabcolsep}{10pt}
\begin{tabular}{l c c c c c}
\toprule
     & General (2D) & Glycine (2D) & Proline (2D) & Pre-Pro (2D) & RNA (7D) \\
     Dataset size & 138208 & 13283 & 7634 & 6910 & 9478 \\
\midrule
    MoPS~\citep{de2020power} & 1.15~\scriptsize{$\pm$ 0.002} & 2.08~\scriptsize{$\pm$ 0.009} & 0.27~\scriptsize{$\pm$ 0.008} & 1.34~\scriptsize{$\pm$ 0.019} & 4.08~\scriptsize{$\pm$ 0.368} \\
    RDM~\citep{huang2022rdm} & 1.04~\scriptsize{$\pm$ 0.012} & 1.97~\scriptsize{$\pm$ 0.012} & \textbf{0.12}~\scriptsize{$\pm$ 0.011} & 1.24~\scriptsize{$\pm$ 0.004} & -3.70~\scriptsize{$\pm$ 0.592} \\
    RFM~\citep{chen2024rfm} & \textbf{1.01}~\scriptsize{$\pm$ 0.025} & \textbf{1.90}~\scriptsize{$\pm$ 0.055} & 0.15~\scriptsize{$\pm$ 0.027} & \textbf{1.18}~\scriptsize{$\pm$ 0.055} & \textbf{-5.20}~\scriptsize{$\pm$ 0.067} \\
\midrule
    Ours (LogBM) & \textbf{1.01}~\scriptsize{$\pm$ 0.026} & \textbf{1.89}~\scriptsize{$\pm$ 0.056} & \textbf{0.14}~\scriptsize{$\pm$ 0.027} & \textbf{1.18}~\scriptsize{$\pm$ 0.059} & \textbf{-5.27}~\scriptsize{$\pm$ 0.090} \\
\bottomrule
\end{tabular}}
\end{table*}
%%%%%%%%%%%%%%%%%%%%%%%%%%%%%%%%%%%%%%%%%
\begin{figure}[!th]
    \centering
\vspace{-0.1in}
    \includegraphics[width=0.8\linewidth]{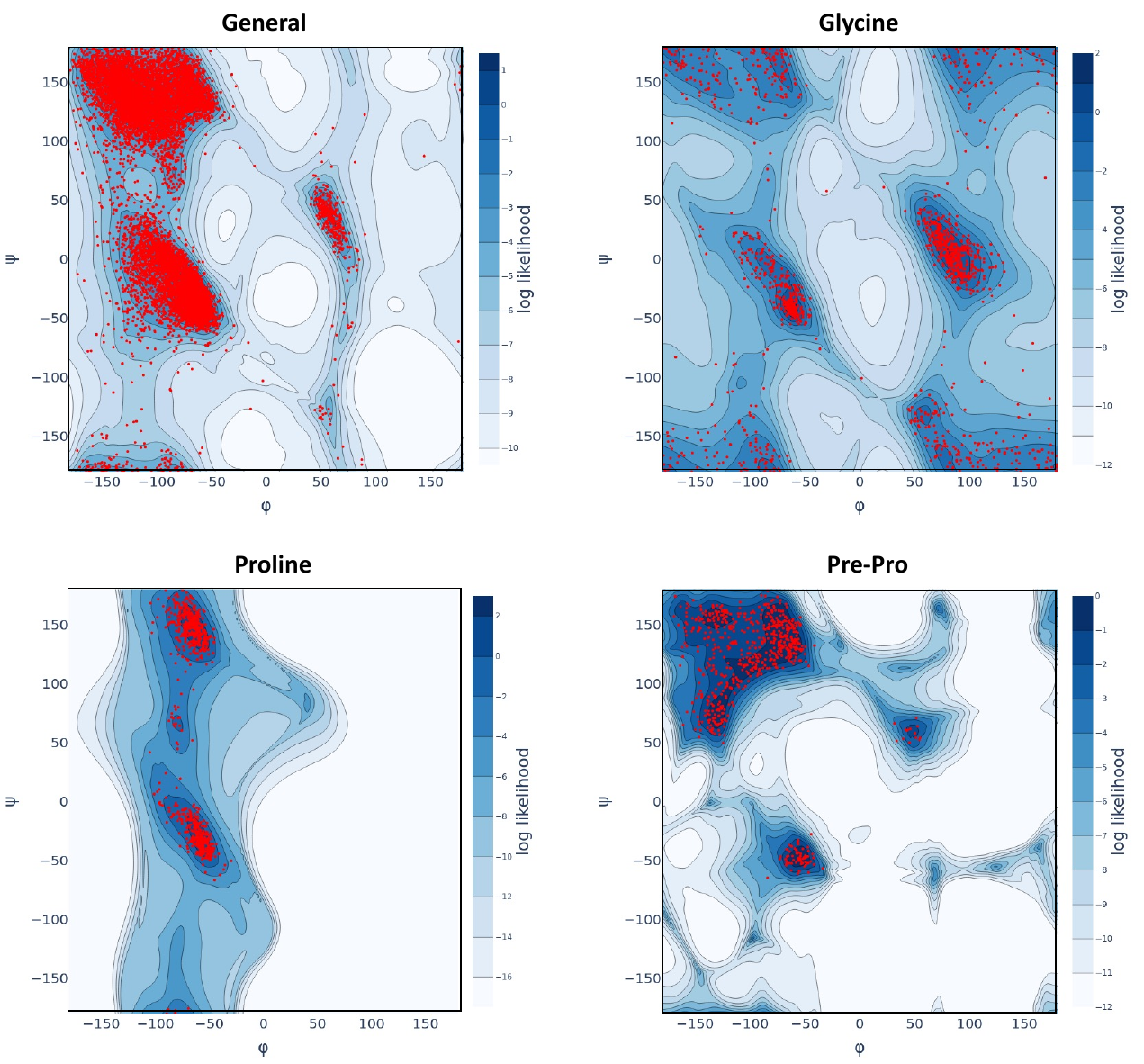}
\vspace{-0.15in}
    \caption{\textbf{Visualization of the learned density} of our model on protein datasets using the Ramachandran contour plots. The red dots denote the samples from the test set. The blue color denotes the log-likelihood computed from our model where the darker colors indicate a higher likelihood.}
\label{fig:ramachandran}
\end{figure}
%%%%%%%%%%%%%%%%%%%%%%%%%%%%%%%%%%%%%%%%%

\section{Experimental Details \label{app:exp}}

\subsection{Implementation Details}
We follow the experimental settings of previous works~\citep{debortoli2022rsgm, chen2024rfm} including the data splits with the same seed values of 0-4 for five different runs. We split the datasets into training, validation, and test sets with (0.8, 0.1, 0.1) proportions. Following \citet{chen2024rfm}, we use the validation NLL for early stopping and the test NLL is computed from the checkpoint that achieved the best validation NLL. 

We parameterize the drifts of the mixture processes with multilayer perceptrons where we concatenate the time to the input, following the previous works. For all experiments except the high dimensional tori, we use 512 hidden units and select the number of layers from 6 to 13, using either the sinusoidal or swish activation function. All models are trained with Adam optimizer and we either do not use a learning rate scheduler or use the scheduler with the learning rate annealed by a linear map which then applies cosine scheduler, as introduced in \citet{debortoli2022rsgm}.
We also use the exponential moving average for the model weights~\cite{polyak1992acceleration} with decay 0.999.

The drifts of the mixture processes are parameterized in the ambient space with projection onto the tangent space as follows:
\begin{align}
    \bm{s}^{\theta}(x, t) = \text{proj}_{x}(\tilde{\bm{s}}^{\theta}(x, t)).
\end{align}
where $\text{proj}_{x}$ is a orthogonal projection onto the tangent space at $x$.
For all experiments, we train our models using the time-scaled two-way bridge matching in Eq.~\eqref{eq:importance_sampling}, where we use 15 steps for the in-training simulation carried out by Geodesic Random Walk~\citep{jorgensen1975central, debortoli2022rsgm}. 

Except for the mesh experiments, we compute the likelihood of our parameterized probability flow ODE using Dormand-Prince solver~\citep{dormand1980family} with absolute and relative tolerance of $1e-5$, following the previous works~\citep{debortoli2022rsgm, chen2024rfm}.
For the mesh experiments, we compute the likelihood with 1000 Euler steps with projection after every step as done in \citet{chen2024rfm}.
For all experiments, we use NVIDIA GeForce RTX 3090 and 2080 Ti and implement the source code with PyTorch~\citep{paszke2019pytorch} and JAX.

\subsection{Earth and Climate Science Datasets}
We follow the data splits of previous works~\citep{debortoli2022rsgm, chen2024rfm}, reporting an average of five different runs with different data splits using the same seed values of 0-4. For a fair comparison with baselines, we set the prior distribution to be a uniform distribution on the sphere.
The convergence analysis demonstrated in Figure~\ref{fig:earth_conv} was conducted on the models trained on the Volcano dataset, where we measure the geodesic distance between the final sample and the trajectory $\bm{Z}_t$ of each method for discretized time steps $t$. The convergence of the predictions was also measured similarly, where we use the parameterized prediction of Eq.~\eqref{eq:prediction}.

\subsection{Protein Datasets \label{app:exp:protein}}
We follow the experimental setup of \citet{huang2022rdm} and \citet{chen2024rfm} where we use the dataset compiled by \citet{huang2022rdm} that consists of 500 high-resolution proteins~\citep{lovell2003structure} and 113 selected RNA sequences~\cite{murray2003rna}. The proteins and the RNAs are divided into monomers where the joint structures are removed and use the backbone conformation of the monomer. For proteins, this results in 3 torsion angles of the amino acid where the 180$^{\circ}$ angle is removed and can be represented on the 2D torus. For RNAs, the 7 torsion angles are represented on the 7D torus. 
We follow the data splits of \citet{chen2024rfm} and report an average of five different runs with different data splits using the same seed values of 0-4. For a fair comparison with the baselines, we also set the prior distribution to be a uniform distribution on the 2D and 7D tori.

%%%%%%%%%%%%%%%%%%%%%%%%%%%%%%%%%%%%%%%%%
\begin{figure}[!t]
\centering
    \caption{\textbf{Visualization of the generated samples and the learned density} of our method and RFM on the mesh datasets. Blue dots represent the generated samples and darker red colors indicate higher likelihood. The numbers in the parentheses denote the number of in-training simulation steps used to train the model.}
\vspace{-0.1in}
    \includegraphics[width=1\linewidth]{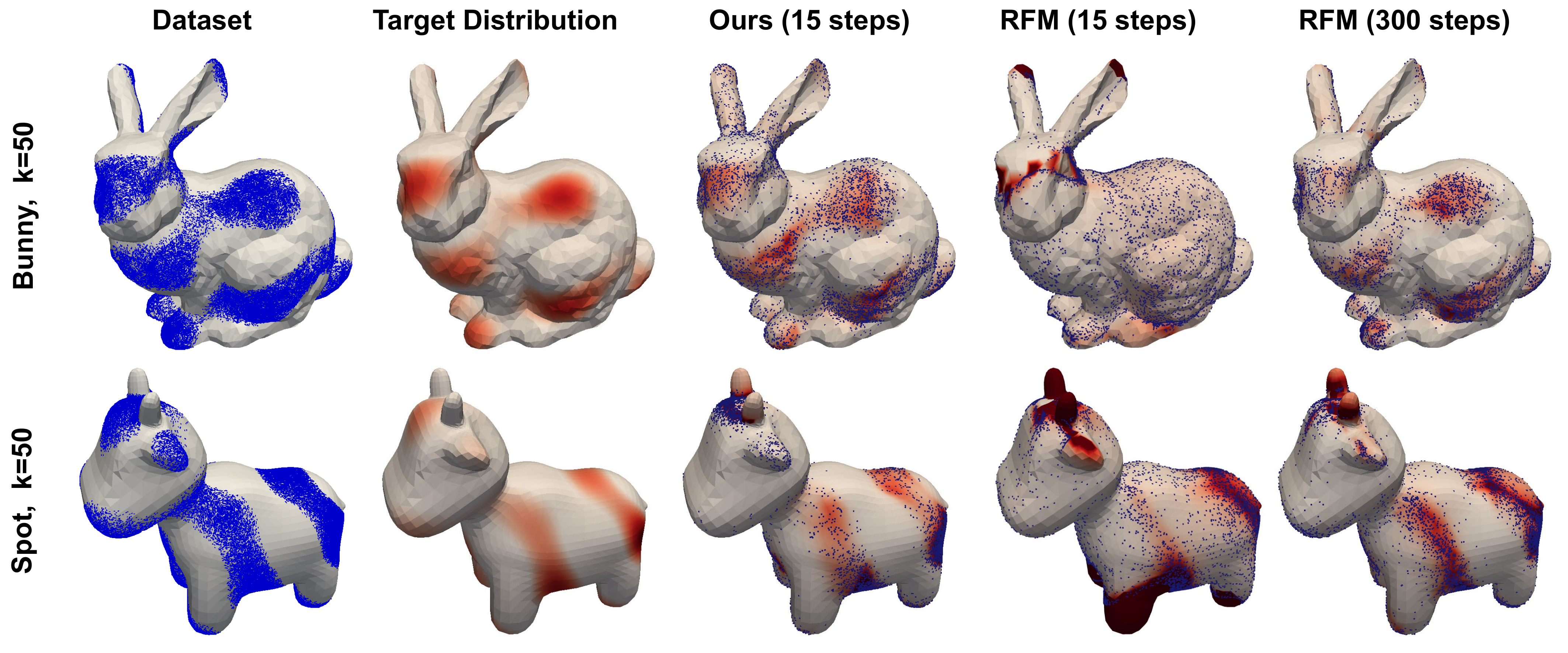}
\label{fig:mesh_vis_abl}
\vspace{-0.2in}
\end{figure}
%%%%%%%%%%%%%%%%%%%%%%%%%%%%%%%%%%%%%%%%%

\subsection{High-Dimensional Tori \label{app:exp:htori}}
We follow the experimental setup of \citet{debortoli2022rsgm} where we create the dataset as a wrapped Gaussian distribution on a high dimensional tori with uniformly sampled mean and scale of 0.2. Since we evaluate on higher dimensions, up to 2000 dimensions, we use 2048 hidden units for all methods. Specifically, we use MLP with 3 hidden layers and 2048 hidden units for RSGM~\citep{debortoli2022rsgm} and RFM~\citep{chen2024rfm}.
To make a fair comparison with the baselines, we match the number of model parameters by using MLP with 2 hidden layers and 2048 hidden units for the model estimating the mixture process, i.e. $\bm{s}^{\theta}_f$, and use MLP with 1 hidden layer and 512 hidden units for the model estimating the time-reversed mixture process, i.e. $\bm{s}^{\phi}_b$. 
We train all methods for 50k iterations with a batch size of 512 without early stopping and evaluate the log-likelihood per dimension for 20k generated samples. We also set the prior distribution to be a uniform distribution on the high-dimensional tori. We measure the training time of each method implemented by JAX~\citep{bradbury2018jax} for a fair comparison.

\subsection{General Closed Manifolds \label{app:exp:mesh}}
We use the triangular meshes provided by \citet{chen2024rfm}: An open-source mesh is used for Spot the Cow and a downsampled mesh with 5000 triangles is used for Stanford Bunny, where the 3D coordinates of the meshes are normalized so that the points lie in $(0,1)$. Following \citet{chen2024rfm}, the target distributions on the mesh are created by first computing the $k$-th eigenfunction (associated with non-zero eigenvalue) on three times upsampled mesh, thresholding at zero, and then normalizing the resulting function. The visualization of generated samples and learned density of RFM in Figure~\ref{fig:mesh_vis} and \ref{fig:mesh_vis_abl} are obtained by running the open source code. For a fair comparison with RFM, we set the prior distribution to be a uniform distribution on the mesh. We measure the training time of our model and RFM which are all implemented in JAX for a fair comparison.

\subsection{Further Analysis}

\paragraph{Time-scaled Training Objective}
We compare our framework trained with Eq.~\eqref{eq:importance_sampling} against a variant trained with uniformly distributed time instead of time-scaled distribution $q$ in Eq.~\eqref{eq:importance_sampling}, on the Volcano dataset. We follow the experimental setup of earth and climate science experiments.

\begin{figure}[!t]
    \centering
    \includegraphics[width=0.98\linewidth]{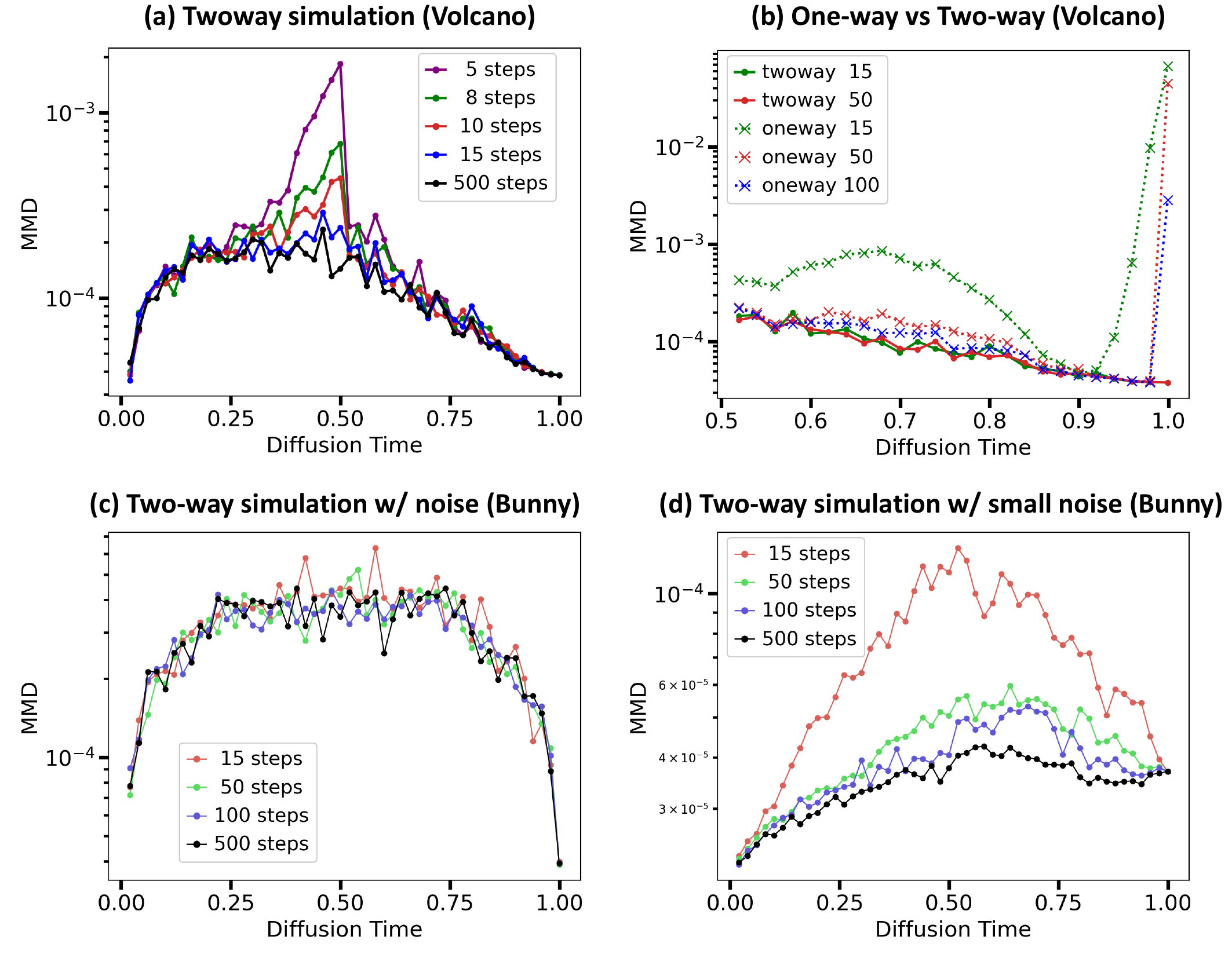}
\vspace{-0.15in}
    \caption{Denoting $\bm{X}_t^{(N)}$ as $\bm{X}_t$ of LogBM simulated with N steps, we measure the \textbf{MMD distance} between $\bm{X}_t^{(500)}$, i.e., almost exact sample, and $\bm{X}_t^{(N)}$ for $N\leq 500$.
    \textbf{(a) Results by differing the number of steps for the two-way approach} where we observe that the MMD results of 15 steps are almost the same as the exact simulation. 
    \textbf{(b) Results with the one-way approach} where we can see that the one-way approach requires a significantly large number of steps to obtain accurate trajectories. 
    \textbf{(c) Results for simulating the mixture process} on the Stanford Bunny dataset where we observe that 15 steps are enough to obtain accurate trajectories. 
    \textbf{(d) Results for simulating the mixture process with small noise scale} where using 15 steps produces highly inaccurate trajectories, and we can see that it requires more than 100 steps to obtain accurate trajectories.}
\label{fig:abl_num_steps}
\end{figure}

\begin{figure}[!t]
    \centering
    \includegraphics[width=0.98\linewidth]{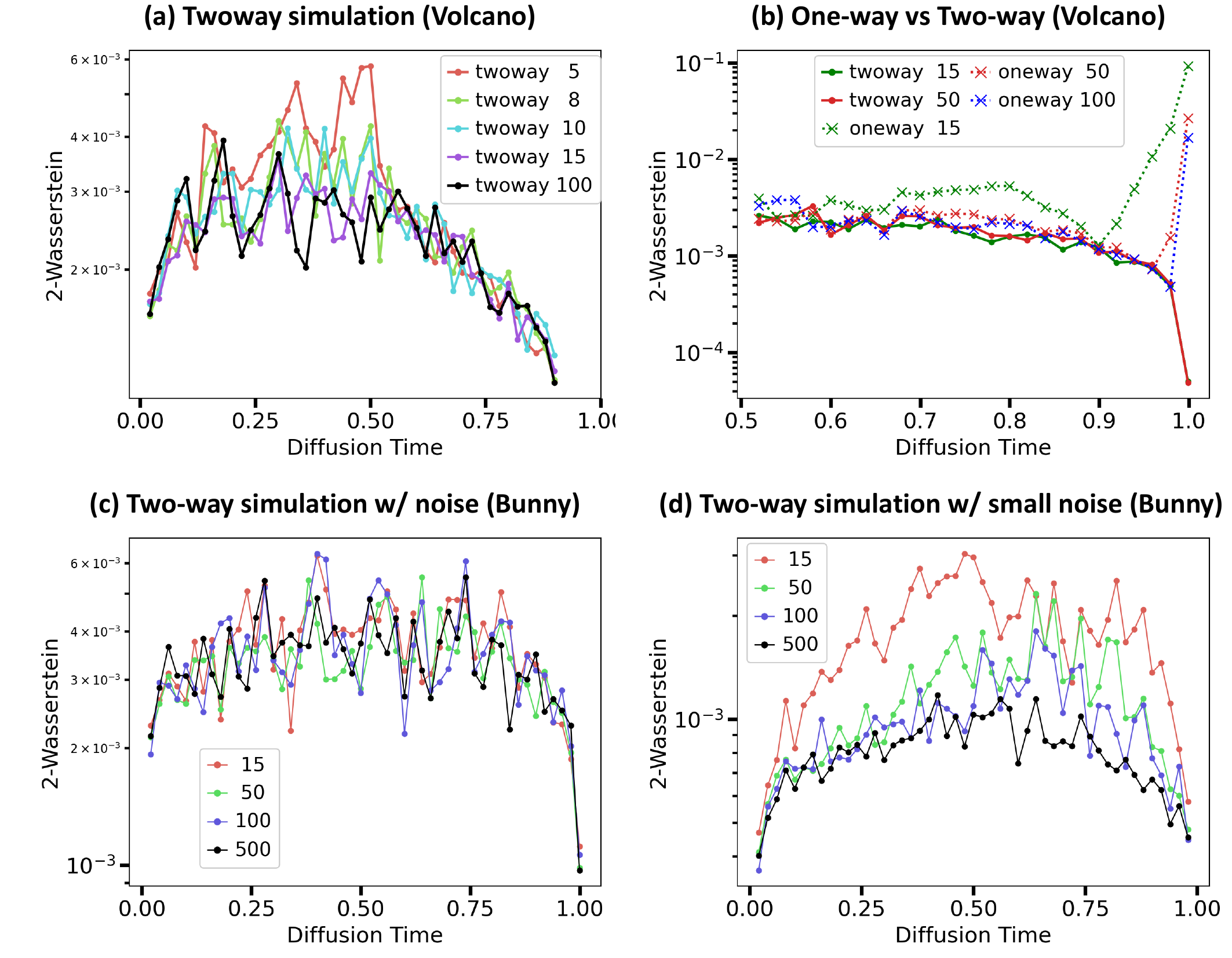}
\vspace{-0.15in}
    \caption{Denoting $\bm{X}_t^{(N)}$ as $\bm{X}_t$ of LogBM simulated with N steps, we measure the \textbf{Wasserstein distance} between $\bm{X}_t^{(500)}$, i.e., almost exact sample, and $\bm{X}_t^{(N)}$ for $N\leq 500$.
    \textbf{(a) Results by differing the number of steps for the two-way approach} where we observe that the MMD results of 15 steps are almost the same as the exact simulation. 
    \textbf{(b) Results with the one-way approach} where we can see that the one-way approach requires a significantly large number of steps to obtain accurate trajectories. 
    \textbf{(c) Results for simulating the mixture process} on the Stanford Bunny dataset where we observe that 15 steps are enough to obtain accurate trajectories. 
    \textbf{(d) Results for simulating the mixture process with small noise scale} where using 15 steps produces highly inaccurate trajectories, and we can see that it requires more than 100 steps to obtain accurate trajectories.}
\label{fig:abl_num_steps_w}
\end{figure}

\paragraph{Number of In-Training Simulation Steps}
For $t$ in $[0,1]$, let $\bm{X}_t^{(N)}$ be the sample $\bm{X}_t$ obtained by simulating LogBM with $N$ discretized steps either from time $0$ to $t$ or $T$ to $t$.
We measure the maximum mean discrepancy (MMD)~\citep{gretton2012kernel} and Wasserstein distance between $\bm{X}_t^{(500)}$ and $\bm{X}_t^{(N)}$ for $N\leq 500$, where $\bm{X}_t^{(500)}$.
For Figures~\ref{fig:abl_num_steps} and \ref{fig:abl_num_steps_w} (d), we use a very small noise scale for the mixture process to mimic a deterministic process. Note that the absolute scales of the MMD among Figure~\ref{fig:abl_num_steps} (a)-(d) are not directly comparable, as the MMD are measured for different reference distributions. The MMD results should be interpreted as how much they deviate from the MMD result of the 'almost exact' trajectories. 

%%%%%%%%%%%%%%%%%%%%%%%%%%%%%%%%%%%%%%%%%%%%%%%%%
\begin{figure*}[!t]
\vspace{-0.3in}
    \centering
    \includegraphics[width=1\linewidth]{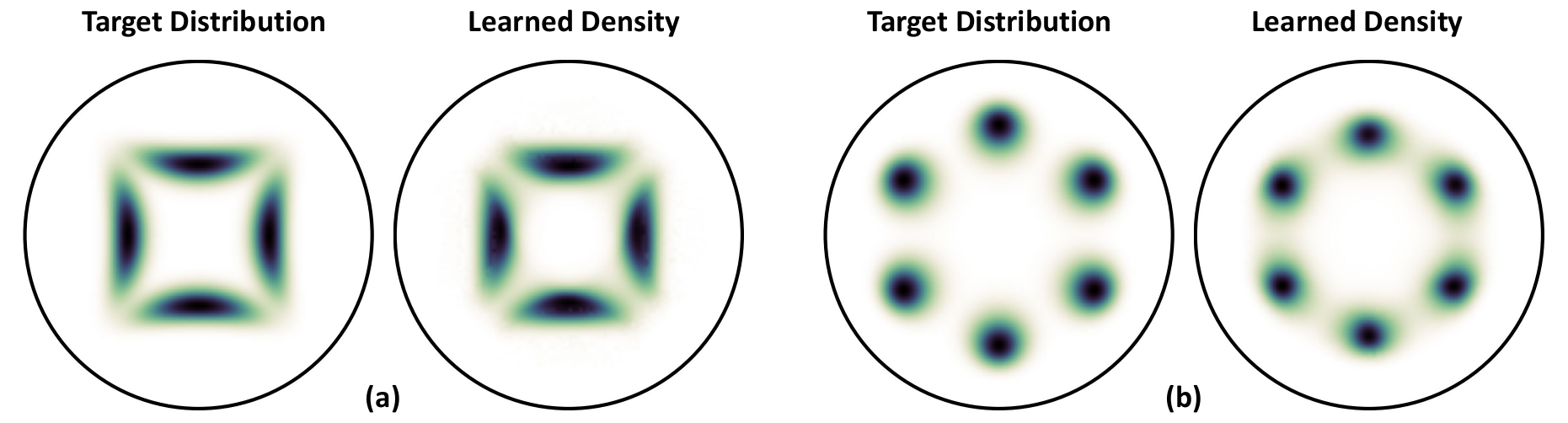}
\vspace{-0.2in}
    \caption{\textbf{Visualization of the learned density} of our model on the hyperboloid. We visualize the synthetic distributions and the learned density on the hyperboloid via projection onto a Poincare disk. (a) visualizes a mixture of four wrapped Gaussian distributions and (b) visualizes a mixture of six wrapped Gaussian distributions.}
\label{fig:hyperboloid}
\vspace{-0.1in}
\end{figure*}
%%%%%%%%%%%%%%%%%%%%%%%%%%%%%%%%%%%%%%%%%%%%%%%%%

\paragraph{Non-Compact Manifold}
We create the synthetic distributions on a 2-dimensional hyperboloid using a mixture of wrapped Gaussian distributions. We use MLP with 4 layers with 512 hidden units and trained for 100k iterations without early stopping. We visualize the learned density in Figure~\ref{fig:hyperboloid} by projecting onto a Poincare disk.

%%%%%%%%%%%%%%%%%%%%%%%%%%%%%%%%%%%%%%%%%
%%%%%%%%%%%%%%%%%%%%%%%%%%%%%%%%%%%%%%%%%

%%%%%%%%%%%%%%%%%%%%%%%%%%%%%%%%%%%%%%%%%
%%%%%%%%%%%%%%%%%%%%%%%%%%%%%%%%%%%%%%%%%

\section{Limitations}
While our method shows superior performance on diverse manifolds, the theoretical guarantee for constructing the mixture process on non-compact manifolds is not sufficient. We experimentally show that our framework is capable of modeling distributions on non-compact manifold with negative curvature.
Also, in our work, we validate that our approach can scale to higher dimensions on manifolds that are considered for real-world tasks, e.g., hypersphere and high-dimensional torus. 
Yet our method may have difficulty when scaling to complex manifolds for which computing the logarithm map or the eigenfunctions are expensive.

\end{document}